\documentclass[english,twocolumn,transaction,10pt]{IEEEtran}

\usepackage[T1]{fontenc}
\usepackage{amsmath,amsfonts,amssymb}
\usepackage{graphicx}
\usepackage{url}
\usepackage{cite}
\usepackage{textcomp}
\usepackage{xcolor}
\usepackage{verbatim}
\usepackage{comment}
\usepackage{bm}

\usepackage[caption=false,font=scriptsize,labelfont=sf,textfont=rm]{subfig}
\usepackage{booktabs} 
\usepackage{array}
\usepackage{stfloats}
\usepackage{makecell}
\usepackage{float}
\usepackage{placeins}
\usepackage{tcolorbox}

\usepackage{algorithm}
\usepackage{algorithmic}

\usepackage{stackrel}
\usepackage{cases}
\usepackage{setspace}

\newtheorem{remark}{Remark}

\begin{document}
	
	\title{Adaptive Dual-Path Framework for Covert Semantic Communication}
	
\author{Xi Yu, Weicai Li,~\IEEEmembership{Graduate Student Member,~IEEE}, Lin Yin, and Tiejun~Lv
		
		\thanks{Manuscript received 2 September 2025; revised 21 January 2026, and 26 March 2026; accepted 27 April 2026. This paper was supported in part by the National Natural Science Foundation of China under No. 62271068. (\emph{corresponding author: Tiejun Lv}.)}
        \thanks{X. Yu, L. Yin, and T. Lv are with the School of Information and Communication Engineering, Beijing University of Posts and Telecommunications (BUPT), Beijing 100876, China (e-mail: \{yusy, yinlin, lvtiejun\}@bupt.edu.cn).}
        \thanks{W. Li is with the Center for Target Cognition Information Processing Science and Technology, and the Key Laboratory of Modern Measurement and Control Technology, Ministry of Education, both at Beijing Information Science and Technology University, Beijing, China (e-mail: liweicai@bupt.edu.cn).}
	}

	\markboth{}%
	{}
	
	\IEEEpubid{}
	
	\maketitle
	
	\begin{abstract}
		This paper proposes a novel adaptive dual-path framework for covert semantic communication (SemCom), which integrates covert information transmission with task-oriented semantic coding. Unlike conventional covert communication methods that embed hidden messages through power-domain signal superposition, our framework embeds covert data within task-specific features via semantic-level intrinsic encoding. This new architecture introduces dual encoding paths with adaptive block selection: an Explicit path for public task execution and a Stego path that jointly encodes both public and covert information through contrastive representation alignment. A Gumbel-Softmax enabled adaptive path selection mechanism dynamically activates network blocks based on task requirements. We formulate a multi-objective optimization framework that simultaneously ensures accurate semantic understanding and reliable covert transmission. 
		We rigorously evaluate our framework's security against a powerful, independently trained attacker. Experimental results on the Cityscapes dataset demonstrate a state-of-the-art level of covertness: our method suppresses the attacker's detection accuracy to a near-random guessing level of 56.12\%. This robust security is achieved while simultaneously maintaining superior performance on the primary semantic tasks compared to the baselines.
	\end{abstract}
	
	\begin{IEEEkeywords}
		Semantic communication, covert communication, physical layer security, contrastive learning, multi-objective optimization.
	\end{IEEEkeywords}
	
	\section{Introduction}
	\IEEEPARstart{I}{n} real-time wireless applications, e.g., the Internet of Things (IoT), autonomous vehicles, and smart cities, semantic communication (SemCom) has emerged as a transformative paradigm to address the dual challenges of data intensity and task complexity~\cite{10183798}.
    By extracting and transmitting only task-relevant features, SemCom eliminates the redundancy inherent in raw data~\cite{9955525}.
    This mitigates key wireless constraints, e.g., limited bandwidth, channel variability, energy efficiency, and latency, by reducing transmission volume, enhancing noise resilience, lowering power consumption, and accelerating task-oriented decision-making~\cite{bourtsoulatze2019deep}.
    Prioritizing semantic content over raw data, SemCom enables efficient, reliable, and low-latency communication in bandwidth-constrained environments, making it a key enabler for next-generation wireless systems, e.g., 5G/6G, edge computing~\cite{10319671}.

\subsection{Status Quo and Research Challenges}
Recognizing transmission efficiency and robustness, recent studies have begun exploring the integration of SemCom into covert communications~\cite{10551876,10090449}. 
These efforts have attempted to exploit the semantic-aware representation of source information to improve the stealth and efficiency of covert transmissions. 

{\color{black}
However, existing approaches typically treat covert communication as an extrinsic data-hiding problem, where an independent secret message is superimposed onto a carrier  (e.g., background noise or interference)~\cite{10095748,10024765,10473138,10328637}. 
This methodology has a fundamental limitation: adding an external signal, even with sophisticated power control, inevitably alters the statistical properties of the original transmission.
While some methods attempt to evade energy-based detectors, this underlying statistical alteration renders them highly vulnerable to modern, trainable attackers capable of identifying non-linear artifacts within the feature space.}

{\color{black}
Beyond extrinsic hiding of an independent secret message,
a parallel branch of research investigates embedding covert information into legitimate public transmissions~\cite{Dutta2013SecretAgent, ArumugamBlochTIFS2019, Liu2017DolphinWhistles}.
Inspired by these embedding-based paradigms, we adopt an intrinsic embedding strategy that utilizes the public information (e.g., semantic segmentation features) as the statistical background for embedding.
However, unlike sacrificial noise carriers, this semantic background provides essential utility to the legitimate receiver.
By exploiting the shared source input, we embed the covert information (e.g., depth features) directly into this functional carrier.
The ultimate goal aligns with the fundamental principle of covertness: ensuring that the joint transmission remains indistinguishable from the benign public-only background, thereby effectively camouflaging the existence of the hidden channel.

To illustrate the practical imperative of this paradigm, consider a secure autonomous driving scenario.
In intelligent transportation networks, standard vehicles broadcast routine semantic segmentation maps for general traffic monitoring (public task), establishing a baseline feature distribution.
However, a specialized vehicle may require the continuous transmission of high-precision depth features for proprietary high-definition map updates (covert task).
If this vehicle were to transmit these dual-task features via explicit dual-task streams, the distinct feature distribution would expose its advanced data processing capabilities.
To prevent this capability exposure, our framework embeds the covert task directly into the segmentation stream.
By strictly aligning the joint feature distribution with that of standard public-only models, the vehicle effectively conceals its advanced capabilities.
This allows it to continuously transmit dual-task data while remaining indistinguishable from ordinary vehicles in the network, thereby achieving capability hiding without sacrificing bandwidth.
}

{\color{black}
However, realizing this imposes a unique challenge that extends beyond standard Multi-task Learning (MTL).
While conventional MTL focuses on maximizing the overall utility of all tasks by learning a shared feature representation, Covert SemCom imposes an additional strict security constraint: the feature representation of the joint task should be indistinguishable to an adversarial detector from that of the single public task.
This creates a fundamental distributional conflict.
Task-specific features for different tasks naturally follow divergent distributions.
Forcing high-capacity covert features into the public stream creates a fidelity-covertness dilemma: optimizing for the covert task tends to distort the public task's feature distribution, producing non-linear artifacts that render the covert transmission highly vulnerable to identification by deep learning-based detectors.
}

{\color{black}  
\subsection{Threat Model for Covert Multi-Task SemCom}
    
    Classical covert communication is primarily concerned with energy-based detection, where an Adversary monitors the physical spectrum to detect power anomalies.
    However, in Covert SemCom, the security battleground shifts to the high-dimensional feature space.
    Since semantic features are abstract representations rather than simple energy pulses, established power-based metrics are insufficient to capture the subtle distributional deviations caused by embedding covert tasks.
    Consequently, to ensure rigorous security assurance, we formulate a threat model that specifically targets these semantic anomalies.

    In this paper, we formalize two distinct classes of Adversarial Detectors to evaluate covertness. 
    The first is a \textbf{Heuristic Detector}, which relies on fixed, predefined statistical metrics (e.g., feature similarity) to identify outliers. 
    The second, and significantly more challenging threat, is the \textbf{Trainable Attacker}. 
    Here, the Adversary is equipped with a deep neural network trained on captured traffic to learn the specific distributional boundaries between public-only transmissions and joint public–covert transmissions~\cite{chapman2018sad}. 
    The security of our framework is rigorously evaluated against this stronger threat to demonstrate robustness against modern, data-driven attacks.
    }

       \subsection{Our Approach and Contributions}

To address the aforementioned threat, we propose a novel adaptive dual-path framework that leverages the architectural properties of multi-task SemCom. Personalized decoders in multi-task SemCom enable a new form of covert communication. Specifically, hidden information can be embedded into public transmissions at the semantic feature level. Only the intended recipient, equipped with the correct personalized decoder, can extract it, while unintended parties remain unaware. 

{\color{black}
To rigorously enforce this security and resolve the fidelity-covertness dilemma, we do not rely on static, over-parameterized models. Instead, we introduce a synergistic mechanism that combines dynamic topology search to identify the optimal embedding subspace with contrastive alignment to strictly enforce indistinguishability.
}

  The contributions of this paper are summarized as follows:

{\color{black}
\begin{itemize}
    \item 
    We propose a novel adaptive dual-path framework for covert semantic communication. This architecture comprises an Explicit path for public-only transmissions and a Stego path for joint public-covert tasks. This structural separation effectively mitigates the feature conflict between high-fidelity semantic transmission and stealthy embedding.
    \item 
    We formulate a joint optimization problem targeting accurate semantic understanding and reliable covert transmission. By employing a Gumbel-Softmax based policy-driven selection strategy, our method enables the system to identify the optimal encoding topology that maximizes concealment, thereby minimizing the structural artifacts inherent in static architectures.
    \item 
    We introduce a security-oriented contrastive alignment mechanism. This objective enforces statistical consistency between the Explicit and Stego features, ensuring that the covert embedding remains imperceptible even against a supervised trainable attacker.
\end{itemize}
}

	Extensive adversarial evaluations rigorously validate our framework's security and performance. Our primary evaluation employs a powerful trainable attacker that achieves a near-perfect 100\% detection accuracy against baseline methods, proving them fundamentally insecure. In contrast, the same attacker is unable to distinguish our proposed method from public traffic, with its detection accuracy consistently remaining near the 50--56\% random-guessing threshold. In-depth ablation studies further confirm that our proposed components, such as contrastive learning and policy-driven path selection, are critical for achieving this robust security. Crucially, this is accomplished while simultaneously outperforming the baselines on key task performance metrics. Collectively, these results demonstrate our framework's ability to establish a new benchmark in balancing high-fidelity task execution with verifiably secure and undetectable covert communication.
	
	The remainder of this paper is organized as follows. The related works are discussed in Section~\ref{sec:related work}, followed by the system model that describes the proposed covert SemCom system.
	In Section~\ref{sec:multi_objective}, we formulate the joint multi-objective problem for covert SemCom. Experimental setup and results are provided in Section~\ref{section:results}, followed by conclusions in Section~\ref{sec:conclusion}.

    \section{Related Work}~\label{sec:related work}
	To the best of our knowledge, no existing studies have attempted to embed covert features into public semantic features while minimizing the semantic similarity between the original public message and its covertly embedded counterpart.

	\subsection{Covert Communication}
	Most research on covert communication mainly focuses on the technical path under the traditional communication paradigm, across multiple domains. In the time domain, random time-slot selection based on early time-hopping significantly improves covert rates, with later works refining timing using information freshness and asynchrony \cite{Lu2023TimeUncertainty,Dani2021Async}. In the frequency domain, spread spectrum and frequency hopping reduce detectability by lowering power density or rapidly changing carrier frequencies \cite{Hu2023Underwater,Du2024Bionic}. The spatial domain leverages multi-antenna beamforming \cite{Zheng2019MIMO}, intelligent reflecting surface (IRS)-assisted transmission \cite{Wang2021IRS}, and directional mmWave/terahertz links to isolate signal leakage. In the power domain, adaptive power control and artificial noise injection are used to conceal transmission activity \cite{Hu2018GreedyRelay,Shu2019DelayFD,Tao2020NOMA}, jointly balancing stealth and communication performance.
	
	In addition, friendly jammers have been employed to obscure transmitters from Attackers. Some studies, e.g., \cite{10288591,Nguyen2021Latency}, have optimized power allocation between the transmitter and jammer to enhance covertness, but overlooked public task delivery, potentially degrading semantic quality due to interference. Similarly, the authors of \cite{Che2023Jamming} demonstrated that concealing signals in noise is more effective than using adversarial jamming, but they also overlooked the transmission of public messages.
	
	However, most traditional covert communication schemes transmit raw data, producing high resource consumption. Additionally, superposition-based methods require extra power for covert signals or jamming, increasing energy overhead.
	
	\subsection{Multi-Task SemCom}
	
	Multi-task SemCom has been explored to improve the efficiency of computation and storage. Early works such as \cite{sem} designed a coarse-to-fine architecture to enhance image reconstruction while supporting multiple tasks through task-specific decoders. TOSC-SR \cite{ima} applied rate-distortion theory to jointly optimize task performance and image quality. In \cite{amu}, a BERT-based system was developed to support various text tasks by extracting semantic features at the transmitter. A scalable system developed in \cite{sca} dynamically adjusted the encoding rate based on feature importance across tasks.
    U-DeepSC \cite{zhang2024unified} further introduced a lightweight vector-based scheme for dynamic adaptation, though it was trained per task rather than in a fully multi-task setting.

{\color{black}
	Although these frameworks enhance efficiency, they are designed for known, public tasks. 
	Crucially, they prioritize joint utility maximization, which inherently causes distributional shifts from the single-task baseline that act as detectable operational signatures.
	The potential of adapting this architecture, particularly the multi-task encoding architectures, to enforce strict statistical alignment for covert communication remains unexplored. 
	Consequently, how to leverage this multi-task structure to conceal a task from an observer is an open question.
}

	\subsection{Secure SemCom}
{\color{black}	Research in Secure SemCom has two main goals: protecting the privacy of known transmissions and concealing a transmission's existence.}	
	One line of work aims to protect privacy by removing or suppressing sensitive information at the source.  For instance, the authors of~\cite{Erdemir2022Privacy} proposed a variational autoencoder-based JSCC framework over a binary symmetric channel. It adopts adversarial training to erase private attributes from the codewords by minimizing their recoverability. However, such methods prioritize privacy at the expense of utility, often resulting in significantly degraded transmission quality for public messages~\cite{Marchioro2020Adversarial,Erdemir2022Privacy}.
	
	Other studies adopt encryption-based schemes to secure private content, typically encrypting either the entire input~\cite{Marchioro2020Adversarial,Xu2021JointEncryption,Tung2022SecureSemantic} or a disentangled private component~\cite{DIBPPJSCC2024}. While these approaches can effectively ensure privacy, they introduce considerable computational overhead and may compromise reconstruction quality and communication efficiency~\cite{Roy2019MaxEntropy,Bloch2021Overview}. For example, DIB-PPJSCC~\cite{DIBPPJSCC2024} encrypts only the private part using AES to reduce overhead. However, the discreteness of codewords hinders gradient backpropagation~\cite{CovertEmbeddingTheory2023}, and the use of fixed-length encryption introduces latency and potential vulnerabilities in long-term transmissions.

	A more relevant line of work seeks to achieve covert SemCom, aiming to ensure that the existence of private transmissions remains undetectable. The authors of~\cite{PowerControl2024} proposed a reinforcement learning-based power control strategy to optimize transmitter and jammer energy under covert constraints. In~\cite{10570800}, a covert and reliable SemCom framework is developed to address semantic privacy in low-SNR and adversarial settings by employing a full-duplex receiver to conceal the entire transmission process.
	{\color{black}However, these pioneering works are limited as they only counter detectors operating in the power domain. This leaves them vulnerable to sophisticated adversaries who can analyze statistical alterations in the semantic feature space.}

	\subsection{Comparison with Existing Studies}
	Unlike existing works~\cite{PowerControl2024,10570800} detecting covert transmission in the power domain, our proposed covert SemCom system can maintain a constant transmission power by embedding covert information at the feature level, effectively preventing eavesdroppers from detecting the presence of private content via power-based detectors. In addition, the learning-based embedding jointly optimizes utility and privacy, ensuring high-quality delivery of both public and covert tasks while safeguarding sensitive information.
	To align our work with the state-of-the-art, where deep learning-based detectors have become the primary threat~\cite{chapman2018sad, lorch2024landscape}, we are the first to rigorously evaluate a covert SemCom system against a powerful, trainable attacker. This establishes a realistic and challenging benchmark for security in this domain.

	\section{System Model}~\label{sec:System Model}
	In this section, we provide an overview of the proposed covert SemCom system and describe the design of the joint source-channel coding framework.

	\subsection{Overview of the Covert SemCom System}
	
	The proposed covert SemCom system enables end-to-end covert task transmission over a broadcast channel originally designated for public tasks. The system consists of a transmitter, Alice, equipped with an encoder for both public and covert tasks; a receiver, Bob, with a dual decoder for extracting both tasks; and an adversarial receiver, Willie, with a public information decoder and a covert task detector.

	Alice extracts and encodes features relevant to the public task while embedding covert information within these features, ensuring no additional communication resources are consumed. Bob’s dual decoder processes the received encoded features to extract both public and covert information. Meanwhile, Willie decodes the received data to obtain public information while simultaneously analyzing the signal to detect the presence of covert transmission.
	
	\begin{figure}[t]
		\centering
		\includegraphics[width=0.8\linewidth]{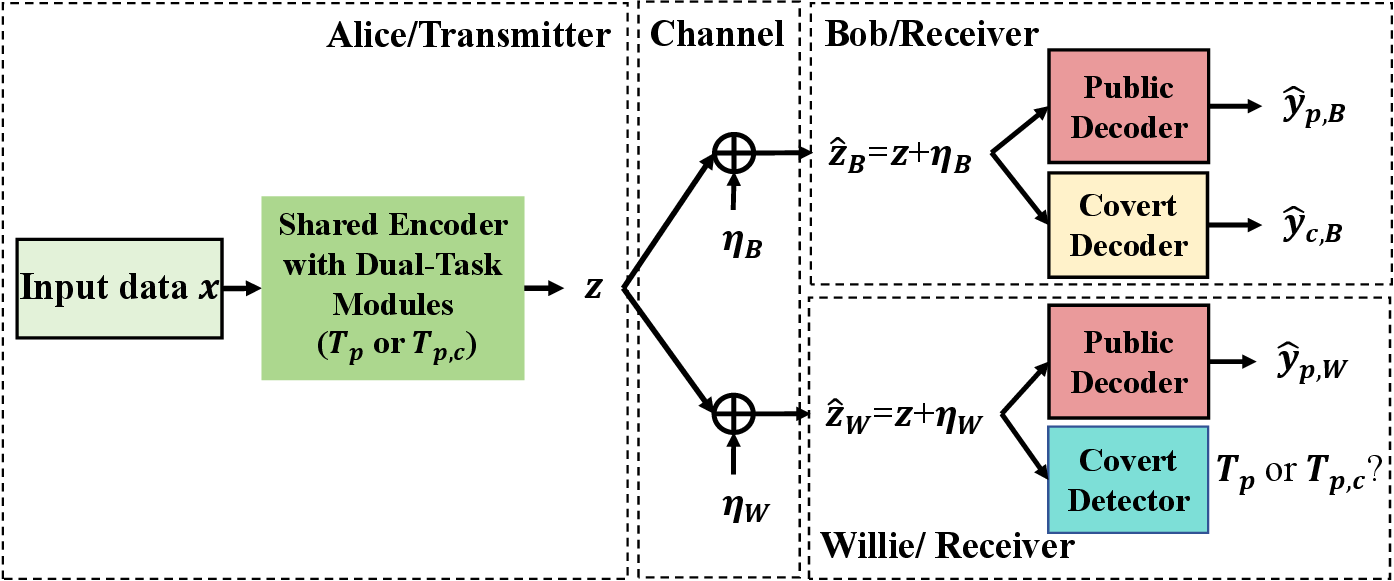}
		\caption{Workflow of the proposed covert SemCom system. The transmitter Alice processes the input image data $\mathbf{x}$ through a Shared Encoder with Dual-Task Modules, selecting between the Explicit path for the public task $T_{\rm p}$ and the Stego path for the joint public/covert tasks $T_{\rm p,c}$ to generate encoded features $\mathbf{z}$. The encoded features are then transmitted over a noisy wireless channel. At the receiver side, Bob receives information and decodes to the public label $\hat{\mathbf{y}}_{\rm p,B}$ and the covert label $\hat{\mathbf{y}}_{\rm c,B}$ using dedicated decoders. Meanwhile, the adversarial receiver Willie receives information and decodes the public label $\hat{\mathbf{y}}_{\rm p,W}$, while simultaneously detecting whether information about the covert task has been transmitted.}
		\label{fig:system_model}
	\end{figure}
	
	The workflow of the proposed covert SemCom system is depicted in Fig. \ref{fig:system_model}.

   {\color{black}
Given the input (image) data \( \mathbf{x} \in \mathbb{R}^{C_{\text{in}} \times  H_{\text{in}}  \times W_{\text{in}}} \) to the encoder, which contains rich semantic information, we focus on the transmission of task-specific semantics. \(C_{\text{in}}\), \(H_{\text{in}}\), and \(W_{\text{in}}\) represent the channel number, height, and width of the input data, respectively.
Let \( \mathbf{x}_{\rm p} \) and \( \mathbf{x}_{\rm c} \) denote the ground truth labels for the public and covert tasks corresponding to the source input \( \mathbf{x} \), respectively.}

	Through the joint source and channel encoder, the input data is compressed into feature \(\mathbf{z} \in \mathbb{R}^{C_{\text{out}} \times  H_{\text{out}} \times W_{\text{out}}}\) for transmitting both public and covert information, where $C_{\rm out}$, $W_{\rm out}$, and $H_{\rm out}$ are the channel number, width, and height of the output, respectively.
	Through the noisy channel, the received information is erroneous, denoted as $\hat{\mathbf{z}}$.
	By decoding the received information $\hat{\mathbf{z}}$, the receiver outputs the tasks \( \hat{\mathbf{y}}_t \), where \( t = \{{\rm p}, {\rm c}\} \) with ``p'' and ``c'' corresponding to the public and covert tasks, respectively.

	Considering channel heterogeneity, the channel conditions for Bob and Willie differ. Bob receives $\hat{\mathbf{z}}_{\rm B} $, and decodes the public task label \( \hat{\mathbf{y}}_{\rm B,p} \) and covert task label \( \hat{\mathbf{y}}_{\rm B,c} \).  Willie receives  $\hat{\mathbf{z}}_{\rm W}$, and decodes the public task label \( \hat{\mathbf{y}}_{\rm W,p} \) while detecting whether covert information is embedded in \( \mathbf{z} \).

	\subsection{Design Details of the JSCC Framework}

	\subsubsection{Encoder with Dual Processing Paths}

    {\color{black}
The core of our framework is an encoder designed to adaptively manage two distinct processing paths: an \textit{Explicit path} for public-only transmissions and a \textit{Stego path} for joint public-covert transmissions. 

To implement this dynamic capability, 
{\color{black}we propose a policy-driven gating mechanism, adapted from the AdaShare framework~\cite{AdaShare}, to selectively share or skip blocks.}
{\color{black}Structurally, this mechanism is built upon a standard CNN backbone (e.g., ResNet) consisting of \(L\) blocks,}
where for each path, a corresponding set of learnable binary selection masks dynamically determines whether a block is executed or skipped. 
This allows our framework to instantiate different computational graphs within a single network, which is essential for balancing task specialization with the efficiency benefits of parameter sharing. 
This adaptive paradigm is critical for our covert framework, as it allows the network to learn two paths that are structurally similar enough to ensure covertness, yet specialized enough to perform their distinct embedding and transmission tasks.

}

	\begin{figure}
		\centering
		\includegraphics[width=1\linewidth]{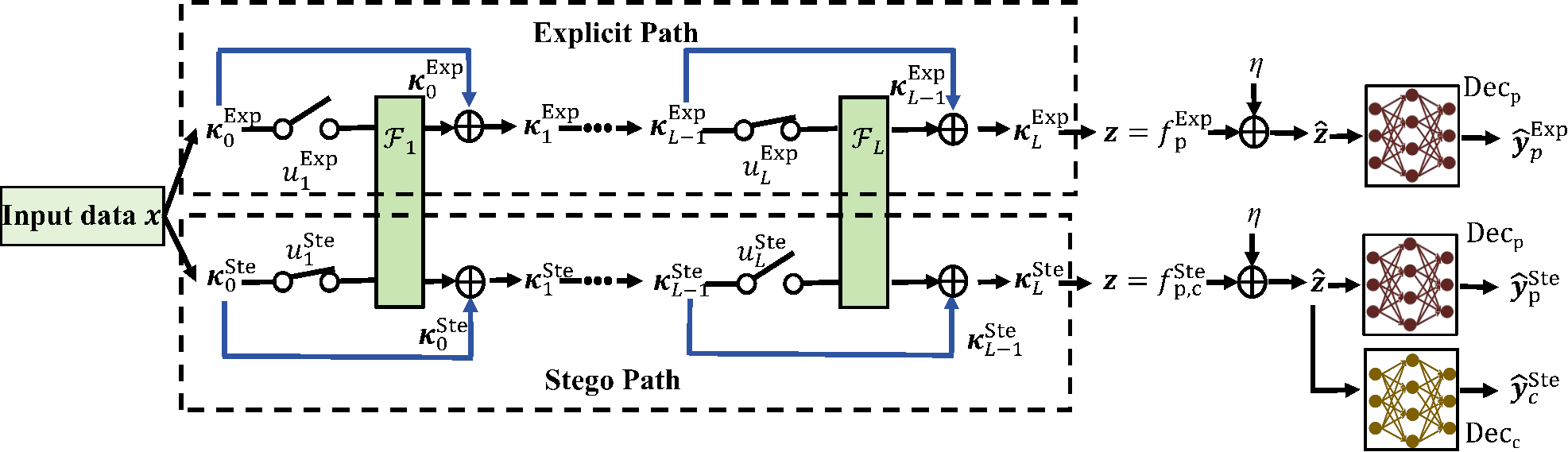}
		\caption{Framework of the encoding and decoding process. The Explicit path is selected to extract the public feature for the public task. The Stego path is pretrained for the joint transmission of the public and covert tasks. The state of the switch, ``open'' ($u_l^{\rm Ste} = 0$) or ``closed'' ($u_l^{\rm Ste} = 1$), influences the selection of the corresponding block.
		}
		\label{fig:Encoding_path}
	\end{figure}
	
	{\color{black}
		\textit{a) Relation to and Novelty over AdaShare:}
		Our work builds upon the adaptive network paradigm introduced by AdaShare~\cite{AdaShare}, a framework originally conceived for the Multi-Task Learning domain. However, we repurpose and fundamentally extend this paradigm to address an entirely different problem space: achieving covert communication under the scrutiny of a third-party observer. This shift from a cooperative MTL setting to a security-sensitive one necessitates a complete redesign of the learning objectives.
		
		Our key methodological novelties begin with a \textbf{new problem formulation} for covert SemCom under a rigorous threat model. To solve this, we introduce a \textbf{security-driven contrastive (CTS) Loss}, a mechanism entirely absent in AdaShare, to enforce statistical indistinguishability against a trainable attacker. Finally, we validate our security claims using a \textbf{rigorous adversarial evaluation framework}, which is standard in the security community but novel to the SemCom context. Thus, our primary contributions are in problem formulation, learning objectives, and security analysis, establishing a new approach for the covert SemCom problem.
	}
	
	\textit{{\color{black}b) Architectural Implementation:}} As shown in Fig.~\ref{fig:Encoding_path}, we need to design two processing paths to meet the task requirements. Since Willie is unaware of the existence of the covert task, Alice, Bob, and Willie jointly train a public encoder and decoder using the Explicit path. Meanwhile, Alice and Bob secretly train another encoding path capable of jointly encoding both public and covert tasks. This Stego path is deployed exclusively for Alice and Bob.

	\begin{itemize}
		\item \textbf{Explicit Path}: For the public task $T_{\rm p}$, we first select an Explicit path to extract the public feature, denoted as $U_{\rm p}^{\rm Exp}=\{u^{\rm Exp}_{ l}|l=1,\dots,L\}$. The output feature of the Explicit path is given by
		\begin{equation}
			f_{\rm p}^{\rm Exp} (\mathbf{x}) =\boldsymbol{\kappa}_{L}^{\rm Exp}\otimes \boldsymbol{\kappa}_{L-1}^{\rm Exp}\otimes\cdots\otimes\boldsymbol{\kappa}_{0}^{\rm Exp},
			\label{eq:Explicit_path}
		\end{equation}
		where $\boldsymbol{\kappa}_0^{\rm Exp} = \mathbf{x}$ is the input data; 
		$\boldsymbol{\kappa}_l^{\rm Exp} = u^{\rm Exp}_{l} \cdot \mathcal{F}_l(\boldsymbol{\kappa}_{l-1}^{\rm Exp})+\boldsymbol{\kappa}_{l-1}^{\rm Exp},l=1,\cdots,L$ denotes the output feature of the $l$-th block; 
		$L$ denotes the total number of blocks of the network;
		$ u^{\rm Exp}_{ l}\in\{0,1\} $  denotes the binary selection of block $l$, which determines whether block $l$ is executed or skipped in the deep neural network when solving public task $T_{\rm p}$;
		$\mathcal{F}_l$  denotes the transfer function of block $l$; \( \otimes \) denotes the block-wise sequential operation that connects the input and output across blocks. 
		
		This Explicit path is jointly trained and shared by Bob and Willie, accompanied by a decoder specifically designed for public information decoding, denoted by $\mathrm{Dec_{\rm p}}$. 
		
\smallskip
        
		\item \textbf{Stego Path}:  To complete the public and covert task, Alice and Bob collaboratively train a Stego path, denoted as $U^{\rm Ste}_{\rm p,c}=\{ u^{\rm Ste}_{ l} |l=1,\dots,L\}$, aiming to embed covert information within the features extracted for the public task. The output feature of the Stego path is given by
		\begin{equation}
			f_{\rm p,c}^{\rm Ste} (\mathbf{x}) =\boldsymbol{\kappa}_{L}^{\rm Ste}\otimes \boldsymbol{\kappa}_{L-1}^{\rm Ste}\otimes\cdots\otimes\boldsymbol{\kappa}_{0}^{\rm Ste},
		\end{equation}
		where $\boldsymbol{\kappa}_0^{\rm Ste} = \mathbf{x}$ is the input data; 
		$\boldsymbol{\kappa}_l^{\rm Ste} = u^{\rm Ste}_{l} \cdot \mathcal{F}_l(\boldsymbol{\kappa}_{l-1}^{\rm Ste})+\boldsymbol{\kappa}_{l-1}^{\rm Ste},l=1,\dots,L$ denotes the output feature of the $l$-th block; the rest of the parameters are consistent with those of the Explicit path.

		This Stego path not only performs the covert transmission between Alice and Bob but also ensures that Bob and Willie can decode the public information using the public decoder $\mathrm{Dec_{\rm p}}$.
		Bob applies a covert decoder $\mathrm{Dec_{\rm c}}$ to recover the covert information.
	\end{itemize}
    
	The covert task is triggered on demand. When no covert task is transmitted, the first path is used for encoding. When covert transmission is required, the second encoder is utilized.
	The output feature for transmission is given by $$\mathbf{z} =
	\begin{cases}
		f_{\rm p}^{\rm Exp} (\mathbf{x}), & \text{when processing } T_{\rm p};\\
		f_{\rm p,c}^{\rm Ste} (\mathbf{x}), & \text{when processing } T_{\rm p} \text{ and } T_{\rm c}.
	\end{cases}$$

	\subsection{Channel Model}
	
	\label{subsec:Channel model}
	
	Consider the application of the proposed JSCC framework over various wireless communication channels, each characterized by distinct statistical models of channel fading. These include commonly adopted models such as the additive white Gaussian noise (AWGN) channel, the Rayleigh fading channel, and the Nakagami-$m$ fading channel, each capturing different aspects of real-world wireless propagation environments.
	
	Summarizing all cases, the received signal is given by 
	\begin{equation}
		\hat{\mathbf{z}} = \boldsymbol{h} \cdot\mathbf{z} + \boldsymbol{\eta},
	\end{equation}
	where $\mathbf{z}$ is the output of the JSCC encoder, $\boldsymbol{h}$ denotes the channel gain, and $\boldsymbol{\eta} \in \mathbb{R}^{C_\text{out} \times H_\text{out} \times W_\text{out}}$ represents the additive noise introduced during noisy channel, which collects the receiver noise with each element sampled from $\mathcal{N}(0, \sigma^2)$. Here, $\mathcal{N}(0, \sigma^2)$ denotes the zero-mean Gaussian distribution with variance~$\sigma^2$. 
	
	For the AWGN channel, we assume a constant channel gain $\boldsymbol{h} = 1$ and Gaussian noise $\boldsymbol{\eta} \sim \mathcal{N}(0, \sigma^2 \mathbf{I})$. In the Rayleigh fading model, $\boldsymbol{h}$ is modeled as a random variable with entries drawn from a circularly symmetric complex Gaussian distribution, i.e., $h_{i} \sim \mathcal{CN}(0,1)$, which reflects non-line-of-sight (NLoS) conditions. For the more general Nakagami-$m$ fading channel, the magnitude $|\boldsymbol{h}|$ follows a Nakagami distribution with shape parameter $m$ and spread parameter $\Omega$, offering a flexible way to model both severe and mild fading scenarios.
	
	These models allow us to comprehensively evaluate the robustness and adaptability of the proposed JSCC framework under diverse wireless channel conditions.

	\subsection{Decoding and Detection}
	In our proposed covert semantic framework, the decoder for the public task, i.e., $\mathrm{Dec}_{\rm p}$, is designed to decode the output from both the Explicit and Stego encoding paths. This ensures that, regardless of whether covert information is embedded, the public task can be successfully decoded without requiring any modification to the decoder at Willie's side. 
	As for the covert task, Bob is equipped with a pre-trained decoder $\mathrm{Dec}_{\rm c}$, while Willie deploys a detector $\mathrm{Det}_{\rm c}$ to identify the presence of covert information. The architectures and implementation details of the dual decoders and the detector are as follows.

	\subsubsection{Design of Public Decoder, $\mathrm{Dec}_{\rm p}$}
	
	Let the received signal $\hat{\mathbf{z}}
	\in\{\hat{\mathbf{z}}_{\rm B},\hat{\mathbf{z}}_{\rm W}\}$ from either the Explicit or Stego path denote as the input of the decoded representation and outputs the public semantic content $\hat{\mathbf{y}}_{\rm p}\in\{\hat{\mathbf{y}}_{\rm p,B},\hat{\mathbf{y}}_{\rm p,W}\}$, denoted as 
    {\color{black}
	\begin{align}
		\hat{\mathbf{y}}_{\rm p}=\begin{cases}
			\hat{\mathbf{y}}_{\rm p}^{\rm Exp}=   \mathrm{Dec}_{\rm p}(\hat{\mathbf{z}}),& \text{when } \mathbf{z}=f_{\rm p}^{\rm Exp} (\mathbf{x});\\
			\hat{\mathbf{y}}_{\rm p}^{\rm Ste}=   \mathrm{Dec}_{\rm p}(\hat{\mathbf{z}}),& \text{when }\mathbf{z}=f_{\rm p,c}^{\rm Ste} (\mathbf{x}).
		\end{cases}
	\end{align}
    }

	The decoder is trained to minimize the loss between the ground truth of the public task $\mathbf{x}_{\rm p}$, and its reconstruction version $\hat{\mathbf{y}}_{\rm p}^{\rm Exp}$, where the encoding process operates solely through the Explicit path. The standard cross-entropy loss of the decoded information and ground truth is defined as:
	\begin{equation}
		\mathcal{L}_{\rm p}^{\rm Exp} = \mathrm{CE}(\hat{\mathbf{y}}_{\rm p}^{\rm Exp}, \mathbf{x}_{\rm p}),\label{loss_exp_p}
	\end{equation}
	where $\mathbf{x}_{\rm p}$ serves as the ground truth label for the public task. To better transmit the public information, $\mathcal{L}_{\rm p}^{\rm Exp}$ is minimized by optimizing the encoder and decoder of the Explicit path.

	\begin{remark}
		$\mathrm{Dec}_{\rm p}$ is shared across both tasks ($T_{\rm p}$ and $T_{\rm p,c}$) and remains unchanged on the receiver side (e.g., Willie), ensuring consistent public task decoding regardless of whether covert information is embedded. This design enables public task recovery while preserving the transparency of covert communication.
		
		To ensure the transmission utility of the public task when embedding the covert information, the cross-entropy loss of the decoded information $  \hat{\mathbf{y}}_{\rm p}^{\rm Ste}$ and the ground truth when processing $T_{\rm p,c}$ is quantified by
		\begin{equation}
			\mathcal{L}_{\rm p}^{\rm Ste} = \mathrm{CE}(\hat{\mathbf{y}}_{\rm p}^{\rm Ste}, \mathbf{x}_{\rm p}).\label{loss_ste_p}
		\end{equation}
		The encoding process for the public task in the Stego path is designed and trained to minimize the corresponding loss $\mathcal{L}_{\rm p}^{\rm Ste}$.
	\end{remark}

	\subsubsection{Design of Covert Decoder, $\mathrm{Dec}_{\rm c}$}
	For the covert task, the decoder $\mathrm{Dec}_{\rm c}$ is trained to reconstruct the covert information $\hat{\mathbf{y}}_{\rm c}$ from the received signal,
	The ground truth label for the covert task is denoted by $\mathbf{x}_{\rm c}$. 
	The reconstruction is given by:
    {\color{black}
	\begin{equation}
		\hat{\mathbf{y}}_{\rm c}^{\rm Ste} = \mathrm{Dec}_{\rm c}(\hat{\mathbf{z}}),
	\end{equation}
    }
	where $\hat{\mathbf{z}}=\boldsymbol{h} \cdot f_{\rm p,c}^{\rm Ste} (\mathbf{x}) + \boldsymbol{\eta}$ is the received signal.

     Given that depth estimation is a regression task, the covert task loss is defined using the $L_1$ norm to minimize the pixel-wise absolute error:
 \begin{equation}
  \mathcal{L}_{\rm c}^{\rm Ste} = \| \hat{\mathbf{y}}_{\rm c}^{\rm Ste} - \mathbf{x}_{\rm c} \|_1. \label{loss_ste_c}
 \end{equation}
 This formulation encourages the covert decoder to precisely recover the depth details by penalizing the deviation between the estimated and ground-truth depth maps.

\subsubsection{Covert Detector}
{\color{black}Unlike a conventional decoder aiming to recover semantic content, Willie may employ a detector designed to analyze the received features and uncover subtle deviations or statistical patterns that indicate the presence of covertly embedded signals. Since the covert information embedding does not alter the overall signal power, conventional energy-based detectors are no longer effective in this scenario. 

{\color{black}More sophisticated strategies for Willie include computing decision statistics based on heuristic metrics, such as feature similarity between a known public transmission and a suspected covert one (corresponding to the \textbf{heuristic detector} in our threat model). Beyond this, a stronger adversary may deploy a \textbf{trainable attacker}, where Willie leverages captured traffic to train a deep learning-based classifier capable of distinguishing public-only transmissions from joint public–covert ones.} To comprehensively evaluate covertness against different types of detectors, different detection models are considered and systematically assessed in Section~\ref{section:results}.}

\section{Multi-objective Covert SemCom }\label{sec:multi_objective}

This section formalizes the joint multi-objective optimization problem for covert SemCom, which aims to learn effective encoding and decoding strategies while ensuring reliable transmission of both public and covert tasks under quality-of-service (QoS) and stealth constraints.

\subsection{Problem Formulation}
The proposed system has multiple design goals, including accurate semantic understanding, reliable covert message transmission, imperceptibility, and efficient feature representation. We formulate a multi-objective loss function composed of four complementary components. Each objective corresponds to a distinct performance criterion, and together they define a unified optimization framework for training the model.

{\color{black}
\subsubsection{Gumbel-Softmax Sampling for Adaptive Path Selection}
\label{sec:gumbel_softmax}

To enable the joint optimization of network topology and weight parameters, we aim to learn the select-or-skip policies for the Explicit and Stego paths ($k \in \{\rm Exp, Ste\}$).
In our formulation, $u_l^k$ serves as a gating coefficient that determines the activation of the $l$-th block.
However, a primary challenge lies in the fact that direct binary selection is non-differentiable, which hinders the backpropagation of gradients.
To overcome this obstacle, we adopt Gumbel-Softmax sampling~\cite{Gumbel} to relax the discrete $u_l^k$ into a continuous variable, thereby bridging the gap between discrete architecture search and continuous gradient descent.

\textit{a) Differentiable Training via Reparameterization:}
Let $\alpha_{l, j}^k$ be the learnable logit for the $l$-th block of path $k$, corresponding to action $j \in \{0, 1\}$ (where $j=0$ denotes \textit{Skip} and $j=1$ denotes \textit{Execute}).
To facilitate efficient exploration and enable gradient flow, we employ the reparameterization trick. 
We first introduce stochasticity by generating independent Gumbel noise $g_{l, j}^k$:
\begin{equation}
    g_{l, j}^k = -\log(-\log(\epsilon_{l, j}^k)), \quad \text{with } \epsilon_{l, j}^k \sim \text{Uniform}(0, 1).
    \label{eq:gumbel_noise}
\end{equation}
Then, to approximate the categorical distribution in a differentiable manner, the soft gate $u_{l}^k$ (corresponding to the \textit{Execute} probability) is computed by applying the softmax function with a temperature $\tau_g$:
\begin{equation}
    u_{l}^k = \frac{\exp((\alpha_{l, 1}^k + g_{l, 1}^k) / \tau_g)}{\sum_{j \in \{0, 1\}} \exp((\alpha_{l, j}^k + g_{l, j}^k) / \tau_g)}.
    \label{eq:gumbel_softmax}
\end{equation}
During training, $u_l^k$ is a \emph{soft} gate with $u_l^k\in(0,1)$; during inference, we use a \emph{hard} binary instantiation of the same gate with $u_l^k\in\{0,1\}$.
This continuous value is directly applied to the block output in the forward pass, ensuring that gradients can propagate back to update the policy logits $\alpha_{l, j}^k$ during training.

\textit{b) Deterministic Inference via Hard Thresholding:}
While soft sampling is essential for training differentiability, a deterministic and sparse topology is required for consistent deployment.
For simplicity, we use the same notation $u_l^k$: it denotes a \emph{soft} gate during training ($u_l^k\in(0,1)$) and its \emph{hard} binary instantiation during inference ($u_l^k\in\{0,1\}$).
Therefore, during inference, we enforce a stable topology by replacing the stochastic soft sampling with a deterministic hard decision based on the logits:
\begin{equation}
    u_l^k = \mathbb{I}\!\left(\alpha_{l,1}^k \ge \alpha_{l,0}^k\right).
    \label{eq:hard_threshold}
\end{equation}
This operation effectively instantiates the learned policy, restoring the discrete nature ($u_l^k \in \{0, 1\}$) of the selection mechanism, where the block is executed ($u_l^k=1$) only if the logit for execution exceeds that of skipping.
}

\subsubsection{Loss Functions vs. Task Goals}
This soft sampling mechanism, i.e., Gumbel-Softmax sampling, enables the discrete selection process to be optimized using gradient-based methods with respect to the following loss functions.

\smallskip
\textit{a) QoS Guarantees for Public and Covert Tasks:} 
To ensure reliable SemCom, the model is optimized to maintain QoS for both public and covert tasks. This is achieved through carefully designed task-specific loss functions:

\begin{itemize}
    \item \textbf{Public Task Loss ($\mathcal{L}_{\rm p} = \mathcal{L}_{\rm p}^{\rm Exp} + \mathcal{L}_{\rm p}^{\rm Ste}$):} 
    This loss evaluates the semantic reconstruction performance of the public task $T_{\rm p}$ over both Explicit and Stego paths. By minimizing $\mathcal{L}_{\rm p}^{\rm Exp}$ in~\eqref{loss_exp_p} and $\mathcal{L}_{\rm p}^{\rm Ste}$ in~\eqref{loss_ste_p}, the model is encouraged to preserve semantic fidelity during transmission, regardless of which path is taken. This contributes to high-quality communication for the public task under dynamic path conditions.

    \item \textbf{Covert Task Loss ($\mathcal{L}_{\rm c}^{\rm Ste}$):} 
    This loss function supervises the covert task $T_{\rm c}$, which involves embedding secret messages into the Stego features and precisely recovering them at the receiver. Minimizing $\mathcal{L}_{\rm c}^{\rm Ste}$ ensures the embedded covert information remains intact and decodable after transmission, offering reliable and stealthy covert communication.
\end{itemize}

\smallskip
\textit{b) Sparsity-Aware Covert Information Embedding:} 
While the task-specific losses focus on QoS for both public and covert tasks, they do not consider the efficiency of representation. In practice, we construct a compact sub-network for each task, in which redundant blocks are skipped without compromising task performance. For this reason, we introduce a sparsity regularization term that encourages minimal and efficient modifications to the public feature space, enhancing the embedding efficiency of the covert task. This design is advantageous in bandwidth-constrained or low-visibility scenarios, where compact and efficient encoding is critical.

We propose a sparsity loss to promote compactness by penalizing the activation probability of each block in the Stego path. Specifically, the loss is formulated as

\begin{align}
\label{eq:sparsity_loss}
	\mathcal{L}_{\mathrm{sparsity}} = \sum_{l=1}^{L} \biggl[ & \beta \cdot (||u_l^{\rm Ste}||_1 + ||u_l^{\rm Exp}||_1) \nonumber \\
	& + \gamma \cdot \frac{L-l}{L} \cdot \text{KL}(q_l^{\rm Ste} || q_l^{\rm Exp}) \biggr],
\end{align}

where
\begin{itemize}
	\item $\|\cdot\|_1$: The $\ell_1$-norm term promotes sparsity in the selection logits, encouraging fewer blocks to be activated;
    \item $q_l^k$: The probability distribution over actions $j \in \{0, 1\}$ (Skip/Execute) at block $l$, derived from the logits $\alpha_{l,j}^k$;
	{\color{black} \item $\text{KL}(\cdot \| \cdot)$: The layer-weighted KL divergence, which regularizes the selection distribution of the Stego Path to align with that of the Explicit Path. The weighting factor $\frac{L-l}{L}$ places a stronger emphasis on enforcing structural similarity in the bottom layers ($l$ is small), which learn generic features, while allowing for more divergence in the top layers ($l$ is large), where task-specific features for the covert task are processed.}
	\item $\beta$ and $\gamma$: Hyperparameter that controls the  sparsity and distribution alignment.
\end{itemize}

\smallskip
\textit{c) Enhancing Concealment via Contrastive Learning:}
A key challenge in our design lies in effectively selecting and separating the Explicit and Stego feature paths for the public and covert tasks, while ensuring maximal concealment to make covert information difficult to detect. To tackle this, we propose an enhanced contrastive learning strategy that guides the model to learn discriminative yet imperceptible path-specific representations.

To further improve the concealment of the covert task, we formulate a contrastive learning objective that promotes similarity between the encoded features of the Stego and Explicit paths. By aligning the representations extracted from the Stego encoder with those from the Explicit encoder, the model reduces discrepancies that could reveal the presence of hidden information, thereby lowering the risk of covert task detection. This is achieved by minimizing the following contrastive loss:
\begin{align}
	\mathcal{L}_{\text{CTS}} =  -\frac{1}{|\boldsymbol{\xi}|} \sum_{\xi_i \in \boldsymbol{\xi}} \log
	\frac{\exp\left( { \boldsymbol{s}[f_{\rm p}^{\rm Exp}(\xi_i), f_{\rm p,c}^{\rm Ste}(\xi_i)]}/{\tau} \right)}
	{\displaystyle \sum_{\xi_j \in \boldsymbol{\xi}} \exp\left( { \boldsymbol{s}[f_{\rm p}^{\rm Exp}(\xi_i), f_{\rm p,c}^{\rm Ste}(\xi_j)]}/{\tau} \right)}.
\end{align}

Minimizing this contrastive loss $\mathcal{L}_{\text{CTS}}$ enforces a strict semantic alignment between the public and covert representations using a batch-wise discrimination strategy.
Functionally, the Explicit feature $f_{\rm p}^{\rm Exp}(\xi_i)$ serves as the anchor.
The objective maximizes the probability of identifying the corresponding Stego feature $f_{\rm p,c}^{\rm Ste}(\xi_i)$ (positive pair) against all other Stego features $f_{\rm p,c}^{\rm Ste}(\xi_j)$ ($j \neq i$) in the batch (negative pairs).
Mechanistically, the numerator pulls the covert embedding $f_{\rm p,c}^{\rm Ste}(\xi_i)$ into the same semantic subspace as its public counterpart.
Simultaneously, the denominator acts as a normalization factor that pushes apart the anchor from the Stego representations of unrelated instances.
This ensures that the Stego path learns to mimic the precise instance-specific semantics of the Explicit path rather than collapsing into a generic distribution, thereby making the joint features indistinguishable from benign public feature.

\subsubsection{Overall Objective Function}

The unified optimization objective jointly considers both task performance and covert communication constraints. It is formulated as:
\begin{align}
   \label{Overall Objective Function}
	&\min_{\boldsymbol{\theta}} \mathcal{L}_{\text{total}} = \mathcal{L}_{\rm p} + \lambda_{\text{c}}\mathcal{L}_{\rm c}^{\rm Ste} 
	+ \mathcal{L}_{\mathrm{sparsity}} +\lambda_{\text{CTS}}\mathcal{L}_{\text{CTS}},
\end{align}
where $\boldsymbol{\theta} = \{U_{\rm p}^{\rm Exp}, U_{\rm p,c}^{\rm Ste}, \mathcal{F}_l, \text{Dec}_{\rm p}, \text{Dec}_{\rm c} \}$ represents the set of learnable parameters, including the select-or-skip policies for the Explicit and Stego paths, the network blocks $\mathcal{F}_l$, and the decoders for the public and covert tasks.
{\color{black}
Here, $\mathcal{L}_{\rm p}$ measures the semantic performance of the public task.
Crucially, this loss is calculated on the outputs of both the Explicit path and the Stego path.
By minimizing the public task loss for the Stego path, we explicitly enforce that the covert embedding process does not degrade the semantic quality of the public message. This ensures the system maintains its functional utility comparable to the baseline public-only transmission.
}
The balancing coefficients $\lambda_{\rm c}$ and $\lambda_{\rm CTS}$ control the trade-off between task performance and concealment effectiveness.
By jointly optimizing all components, the model is guided to achieve high-quality SemCom for both public and covert tasks while maintaining stealth and efficiency in the covert transmission.

\subsection{Computational Complexity}
For transmitting a single input data sample over the proposed Covert SemCom system, the computational complexity involves feature extraction through the encoder and feature reconstruction through the decoder. 
Specifically, the encoder first processes the input data through $L$ residual blocks, each contributing a computational cost determined by its structure. 
The computational complexity of the $l$-th block is defined as
\[
C_{l} = 2 \times ( K_l \times K_l \times C_{\text{in}_l} \times C_{\text{out}_l} \times H_l \times W_l),
\]  
where $C_{\text{in}_l}$ and $C_{\text{out}_l}$ denote the numbers of input and output channels of the $l$-th block, respectively, $K_l$ denotes the kernel size, and $H_l \times W_l$ represents the spatial dimensions of the feature map at the $l$-th block. 
The selection of each block is influenced by its switch state, which is set to ``open'' ($u_l^{\rm Ste} = 0$, $u_l^{\rm Exp} = 0$) or ``closed'' ($u_l^{\rm Ste} = 1$, $u_l^{\rm Exp} = 1$). The encoder complexity of path $k \in \{\mathrm{Exp}, \mathrm{Ste}\}$ is given by
\[
C_{\text{Encoder}}^{k} = {\sum}_{l=1}^{L} u_l^kC_{l}.
\]  
This dynamic skipping mechanism allows efficient resource allocation while maintaining task-specific feature extraction.  

Subsequently, the decoder transmits and processes the extracted feature representation, which reconstructs the target output through a series of upsampling and convolutional layers. Both public and covert decoders use $M$ parallel decoding modules per path. The decoder complexity of a single module consisting of two blocks for task $k' \in \{\mathrm{p}, \mathrm{c}\}$ is given by
\begin{align}\notag
	C_{\text{Decoder}}^{k'} =& \left( K_{\text{de1}} \times K_{\text{de1}} \times 512 \times 1024 \right. \\&
	+ K_{\text{de2}} \times K_{\text{de2}} \times 1024 \times C_{\text{out}_L}\left.\right) \times H_L \times W_L,\notag
\end{align}
where $C_{\text{out}_L}$ and $H_L\times W_L$ denote the number of output channels and the spatial dimension of the feature map produced by the last ($L$-th) block of the encoder, respectively; $K_{\text{de1}}$ and $K_{\text{de2}}$ denote the kernel sizes of the first and second decoder blocks, respectively.

The overall computational complexity of the proposed Covert SemCom system is obtained by summing the complexities across all paths and tasks: 
\[
C_{\text{Total}} =\sum_{ k \in \{\rm Exp, Ste\}} C_{\text{Encoder}}^k + M\sum_{k'\in\{\rm p ,c\}} C_{\text{decoder}}^{k'}.
\]  
A quantitative analysis and comparison of this complexity against baseline models is provided in Appendix~\ref{sec:appendix_complexity}.

\section{Experiments and Results} \label{section:results}

In this section, we conduct extensive experiments to evaluate the effectiveness of the proposed model under multiple optimization objectives.
This includes comparisons with classical models and ablation studies to analyze the contributions of different components of the proposed model.

\subsection{Experimental Setup}
We validate our experiments using a widely recognized public dataset, i.e., \textbf{CityScapes}~\cite{7780719}, which supports multiple vision tasks. The CityScapes dataset comprises high-resolution street-view images captured in urban environments. In this study, we employ CityScapes to evaluate our model on two tasks: \textit{Semantic Segmentation} and \textit{Depth Prediction}.

\subsubsection{Task Performance Metrics}

In our setting, the public task $T_{\rm p}$ is defined as a Semantic Segmentation task. We adopt the 19-class annotation scheme and follow the official train/test splits.
The dataset is evaluated using two key metrics for Semantic Segmentation: mean intersection over union (\textit{mIoU}) and pixel accuracy (\textit{Pixel Acc}), both of which aim to be as high as possible. The \textit{mIoU} computes the average ratio of the intersection to the union of predicted and ground-truth regions across all classes, reflecting category-level segmentation accuracy. \textit{Pixel Acc} measures the proportion of correctly classified pixels relative to the total number of pixels in the image.

The Depth Prediction task, denoted as $T_{\rm c}$, serves as the covert task in this system. 
The evaluation metrics include absolute error (\textit{abs err}), relative error (\textit{rel err}), and threshold accuracy (\textit{$\delta$}).
Absolute error quantifies the direct discrepancy between the predicted and actual values, while relative error expresses this discrepancy as a proportion of the actual value.
Another key metric is $\delta$, which measures the relative discrepancy between the predicted depth value (${d}$) and the actual depth value ($d_{gt}$).
Specifically, $\delta$ is defined as $\delta=\max\left(\frac{d}{d_{gt}},\frac{d_{gt}}{d}\right)$, and this metric evaluates the proportion of depth estimates where $\delta$ falls below a specific threshold.
The thresholds used in the dataset are set to $1.25$, $1.25^2$, and $1.25^3$.

\subsubsection{Covertness Evaluation Methodology}
\label{sec:covertness_methodology}

To provide a comprehensive and multi-faceted evaluation of our system's covertness, we employ two complementary analytical approaches: a statistical analysis of feature similarity and a direct trainable attacker. Together, these methods provide a holistic assessment of our system's security.

\textit{a) Statistical Similarity Analysis:}
From a statistical perspective, we first utilize a set of heuristic metrics to directly quantify the similarity between the public feature maps and the steganographic feature maps. These include: (1)~\textbf{Feature Cosine Similarity}, which measures the directional alignment of the feature vectors, and (2)~a \textbf{Mutual Information Proxy}, defined as \(1 - \mathcal{L}_{\text{CTS}}\), which reflects the distributional divergence. These metrics provide a direct, white-box measure of how effectively our model achieves its design objective of feature space alignment.

{\color{black}\textit{b) Adversarial Resilience Analysis:}
	Second, from a functional security perspective, we conduct a rigorous adversarial evaluation. We employ a \textbf{trainable attacker}, which is trained on a large, balanced dataset to distinguish between the two classes of transmissions, to explicitly test the system's resilience in a black-box setting. Its final detection accuracy on a held-out test set serves as a practical measure of covertness, where a result approaching the 50\% random-guessing baseline indicates a high degree of security.

	By considering these two perspectives in tandem--the internal statistical alignment and the external adversarial resilience--we form a complete picture of the system's security. The former explains \textit{how} our method achieves concealment at the feature level, while the latter proves \textit{that} this concealment is effective against a dedicated adversary.}

\subsubsection{Model Architecture and Training Details}

{\color{black}The architecture of the proposed covert SemCom system consists of an encoder and a dual-decoder design, while covertness evaluation is conducted using a trainable adversarial detector.}

\textit{a) Encoder:} The encoder is built upon ResNet-34~\cite{He_2016_CVPR} with 16 blocks, a widely adopted backbone for feature extraction. The encoding paths are determined by the selection of specific blocks, following the design principles of AdaShare~\cite{AdaShare}.

\textit{b) Decoder:} {\color{black}The proposed model incorporates a dual-decoder structure with four dedicated decoding blocks to support the public and covert tasks. It consists of a public task decoder, denoted as $\mathrm{Dec}_{\rm p}$, and a covert task decoder, $\mathrm{Dec}_{\rm c}$. The public decoder is designed with the flexibility to process either standalone public features or the combined Stego features, while the covert decoder is specialized to recover the covert task from the Stego features. Architecturally, each decoding block first employs a dilated convolution layer (kernel size $3$, dilation rate \( r \)) to capture contextual information efficiently. This is followed by two $1\times1$ convolutional layers that perform dimensionality reduction and class prediction. These blocks are further enhanced with ReLU activation functions to learn complex patterns and dropout layers for regularization, ensuring the model can handle both tasks effectively while maintaining computational efficiency and robustness.}

{\color{black}\textit{c) Trainable Attacker:}
	To rigorously evaluate the system's security, we designed a powerful \textbf{trainable attacker} to act as the detector for Willie.

    This attacker is implemented as a dedicated 1D Convolutional Neural Network (CNN), chosen for its effectiveness in analyzing spatial patterns within the feature maps. It consists of two sequential Conv1d blocks followed by an MLP head for binary classification. } 
    {\color{black}
Crucially, the training of this attacker is conducted in a supervised manner under a worst-case assumption. 
We assume a knowledgeable attacker who understands the encoding mechanism and has access to the public data distribution. 
Consequently, the attacker can generate a labeled training set by passing sampled images through the known encoder to produce both benign (Explicit path) and covert (Stego path) feature maps. 
These generated features serve as ground-truth samples to supervise the training of the binary classifier. 
The trained attacker is then evaluated on the actual transmission (held-out test set) to quantify detection accuracy~\cite{7932371}.
}

To train the entire system, we use the Adam optimizer~\cite{kingma2014adam} to update the policy distribution parameters and stochastic gradient descent (SGD)~\cite{kiefer1952stochastic} to optimize the network parameters.
The batch size for both optimization processes is fixed at~4. Once the policy training is complete, we sample network architectures from the trained policy for further evaluation and training.
Specifically, we draw 8 distinct network architectures from the policy distribution and subject each sampled network to retraining for a total of 40,000 steps.
This process enables us to evaluate the policy's effectiveness in generating promising network architectures and to fine-tune these architectures to enhance their performance.

{\color{black}The training convergence and stability of our proposed model and the baselines are demonstrated by the error curves provided in Appendix~\ref{sec:appendix_convergence}, Figure~\ref{fig:convergence_curves_proposed}.}

\subsubsection{Baselines}
We evaluate the proposed method against the following baselines:

\begin{itemize}
    \item \textbf{Stacking-based method (Base Model 1)~\cite{DIBPPJSCC2024,wang2025genran,Dutta2013SecretAgent}:} 
    This approach employs a dual-encoder architecture where features for the public task and the information for the covert task are encoded separately. The two resulting feature streams are then fused via simple element-wise addition to form the final representation.

    \item \textbf{Noise-based method (Base Model 2)~\cite{Hu2018GreedyRelay,Shu2019DelayFD,Tao2020NOMA}:} 
    Based on a traditional method of covert communication that adjusts the noise power, we trained an adjustable noise while maximizing the similarity between covert and public information.
\end{itemize}

\subsection{Experimental Results}

\subsubsection{Comparison with Baselines}
\label{Comparison with Baselines}

\begin{figure}[t]
	\centering
    \subfloat[Statistical Analysis of Covertness (Heuristic Metric)]{\includegraphics[width=0.23\textwidth]{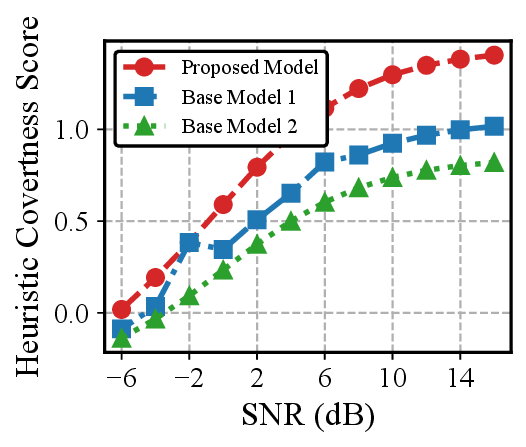}
		\label{fig:covertness_main_results:a}}
	\hfill 
    \subfloat[Adversarial Resilience Analysis (Trainable Attacker)]{\includegraphics[width=0.23\textwidth]{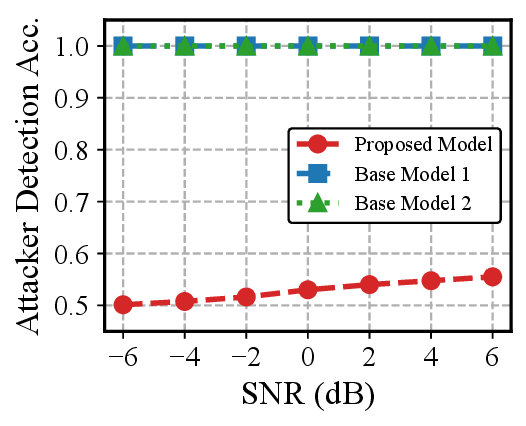}
		\label{fig:covertness_main_results:b}}
	\caption{Comprehensive covertness evaluation under varying SNR conditions. Figure (a) shows the results based on our internal heuristic metric, where higher is better. Figure (b) presents the definitive results from our adversarial evaluation, showing the detection accuracy of a trainable attacker, where lower is better. The stark contrast validates the robustness of our proposed model.}
	\label{fig:covertness_main_results}
\end{figure}

\begin{figure}[t]
	\centering
\includegraphics[width=0.5\textwidth]{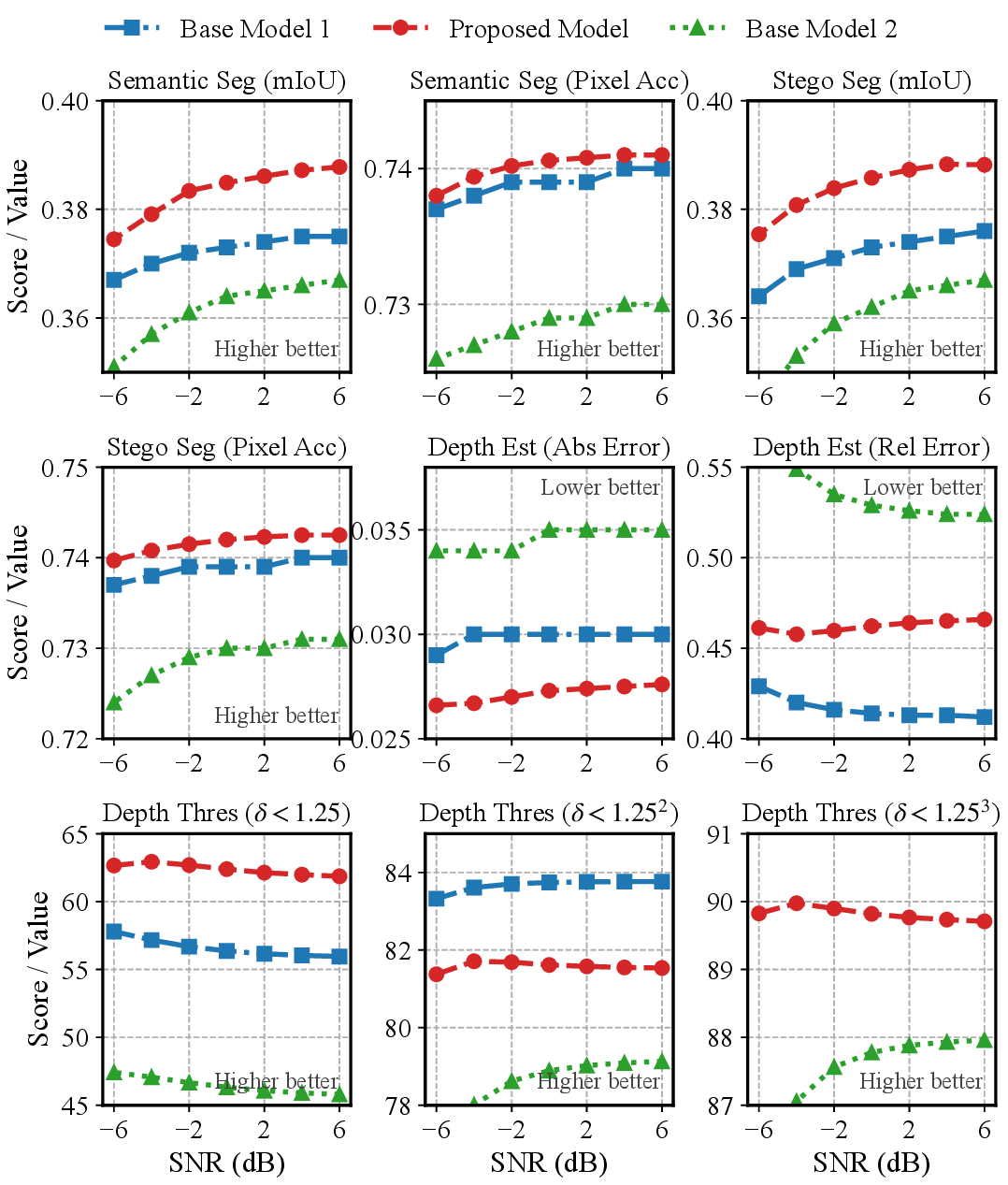}
	\caption{Task performance of different models under varying SNR conditions. Semantic Segmentation is a public task. Depth Prediction is a covert task.}
	\label{fig:task_snr}
\end{figure}

We comprehensively compare our proposed system against two baselines, evaluating both security (Fig.~\ref{fig:covertness_main_results}) and task performance (Fig.~\ref{fig:task_snr}) across a wide range of SNR conditions.

Our security evaluation (Fig.~\ref{fig:covertness_main_results}b) reveals the detection accuracy of a powerful, independently trained attacker. The results reveal a critical security flaw in the baseline methods. The trainable attacker achieves a near-perfect 100\% detection accuracy against both Base Model 1 and Base Model~2, regardless of the channel noise level. This indicates that they possess fundamental structural artifacts that are trivially learned by a dedicated adversary. In stark contrast, our proposed model demonstrates exceptional resilience. The attacker's accuracy against our method remains consistently near the 50\% random-guessing threshold, ranging from a near-perfectly covert 50.12\% at $-6$ dB to a still highly covert 55.50\% at $6$ dB. This provides strong evidence that our method achieves robust covertness through intrinsic statistical indistinguishability. This finding is further corroborated by our internal heuristic metric, shown in Fig.~\ref{fig:covertness_main_results}(a), where our proposed model also consistently outperforms the baselines.

While achieving this state-of-the-art security, our model also excels in task performance, as detailed in Fig.~\ref{fig:task_snr}. Our model consistently outperforms the baselines in most key metrics, including segmentation \textit{mIoU} and pixel accuracy for both the Explicit and Stego paths, as well as depth absolute error. For metrics like \textit{rel err} and the higher-threshold $\delta$ accuracies, the proposed model performs slightly worse. This arises from the contrastive mechanism prioritizing \textit{global feature alignment} to enhance covertness, which may marginally suppress \textit{task-specific fine-grained patterns}. This represents a well-defined trade-off between maximizing security and optimizing for every single task sub-metric. Despite these localized trade-offs, the overall task performance remains superior.

In summary, our comprehensive evaluation demonstrates that the proposed framework is superior in both security and utility. The adversarial analysis proves our system's fundamental security, suppressing a powerful attacker's accuracy to an average of only 52.83\%, while the same attacker achieves 100\% accuracy against baselines. Crucially, this robust security is achieved not by sacrificing, but by simultaneously improving upon the baseline's task performance across most key metrics, establishing our framework's ability to effectively balance the competing demands of utility and stealth.

\begin{figure}[t]
	\centering
    \includegraphics[width=0.5\textwidth]{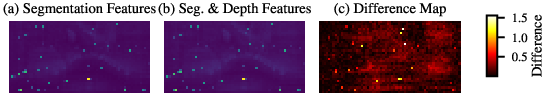}
	\caption{Feature distribution visualization of the proposed model training.}
	\label{fig:feature_cts}
\end{figure}
\begin{figure}[t]
	\centering
    \includegraphics[width=0.5\textwidth]{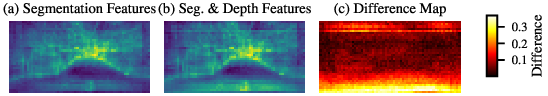}
	\caption{Feature distribution visualization of the Base Model 1.}
	\label{fig:feature_base1}
\end{figure}
\begin{figure}[t]
	\centering
    \includegraphics[width=0.5\textwidth]{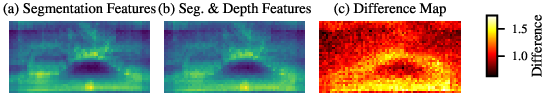}
	\caption{Feature distribution visualization of Base Model 2.}
	\label{fig:feature_base2}
\end{figure}

\subsubsection{Visualization of Features}

To demonstrate the advantages in covertness, we visualize the feature distributions and difference maps of the Explicit Path (encoding for Semantic Segmentation) and Stego Path (encoding for both Semantic Segmentation and Depth Prediction) under 20 dB SNR conditions, as shown in Fig.~\ref{fig:feature_cts} (proposed model), Fig.~\ref{fig:feature_base1} (Base Model 1), and Fig.~\ref{fig:feature_base2} (Base Model 2).
The difference maps are generated by calculating the absolute differences between the features of the two tasks, where the color depth directly reflects the quality of covertness: darker colors (black) indicate smaller differences, tighter integration of public and covert task features, and stronger covertness. 

The difference map for the proposed model is predominantly characterized by black regions, covering over 70\% of the area. This indicates a high degree of alignment between the feature distributions of the public and covert tasks. By enforcing feature space alignment through the contrastive mechanism, covert information is effectively encoded as subtle noise patterns within the public features. Only minor differences, manifested as sparse bright spots, persist in critical regions such as object edges. This ensures robust task performance while avoiding statistical anomalies that could compromise covertness.

In contrast, the difference map of Base Model 1 displays a uniform bright yellow across the entire area, with virtually no significant dark regions. The independent optimization of public and covert features results in a clear separation of feature distributions between tasks. This makes the covert information highly detectable, leading to poor covertness as the model fails to effectively integrate covert features into public representations.

Base Model 2 exhibits large bright white regions covering the majority of the map, interspersed with only sparse dark spots. The presence of extensive bright colored regions with high differential values indicates that the noise injection approach fails to achieve effective feature alignment between tasks. Instead of concealing covert information, this approach introduces additional interference, further highlighting the superiority of the proposed contrastive mechanism in balancing task performance and covertness.

The proposed model's difference map shows a dominant black region coverage ($70\%$ vs. $<10\%$ for the base models), validating its superiority in covertness. Despite minor bright regions (retained for critical task performance), the global feature integration makes it difficult for {\color{black}the detector} to statistically separate covert information. 
Compared to the base models, our model demonstrates robustness to channel noise and achieves better task performance and covert communication performance across the entire SNR range, highlighting its effectiveness in balancing task performance and covert communication under various noise conditions. 

{\color{black}

\subsection{Comparison with Standard SemCom}

To rigorously quantify the performance overhead induced by the proposed covert transmission mechanism, we benchmark our framework against a \textit{Standard SemCom} baseline. Unlike our proposed approach where the covert task is embedded within the public feature stream, this baseline operates under a standard multi-task paradigm: it utilizes the backbone to establish two independent, parallel branches. 
Specifically, each path is explicitly dedicated to transmitting its respective task features (segmentation and depth) without any cross-task embedding or distributional alignment constraints, thereby serving as an unconstrained utility upper bound under the same backbone.

The comparative results are summarized in Table \ref{tab:full_comparison}. Analysis reveals that our framework achieves a highly favorable trade-off. While the unconstrained Standard SemCom naturally achieves higher utility, our method maintains high public fidelity (with a negligible 0.52\% drop in Pixel Accuracy) and preserves the essential structural integrity of the covert task. Although the \textit{abs err} increases slightly (+0.0058), the structural fidelity metric $\delta < 1.25^3$ drops only 3.84\% (relative), indicating that the dense depth semantics are well-retained. Crucially, this moderate utility cost acts as a necessary premium for security, successfully suppressing the adversarial detection accuracy from 100\% (complete exposure) to 56.12\% (robust covertness). Standard SemCom fails because it generates covert features independently. Lacking feature embedding and alignment, it produces distinct statistical artifacts that are trivial to detect. Furthermore, as detailed in Section~\ref{Comparison with Baselines} (Fig.~\ref{fig:covertness_main_results} and Fig.~\ref{fig:task_snr}), our approach significantly outperforms other covert baselines, confirming that it occupies the optimal operating point between theoretical utility and practical security.

\begin{table}[ht]
    \scriptsize
    \color{black}
    \centering
    \caption{Comparison with Standard SemCom.}
    \label{tab:full_comparison}
    \setlength{\tabcolsep}{0.02cm} 
    \begin{tabular}{l|cc|ccc|cc}
        \toprule 
        & \multicolumn{2}{c}{\textbf{Public Task}}
        & \multicolumn{3}{c}{\textbf{Covert Task (Depth)}}
        & \multicolumn{2}{c}{\textbf{Security Metrics}}
        \\
        \cmidrule(lr){2-3} \cmidrule(lr){4-6} \cmidrule(lr){7-8}
        
        \textbf{Method}
        & \makecell{\textbf{mIoU} $\uparrow$} 
        & \makecell{\textbf{Pixel} \\ \textbf{Acc} $\uparrow$} 
        & \makecell{\textbf{abs} \\ \textbf{err} $\downarrow$} 
        & \makecell{\textbf{$\delta <$} \\ 1.25 $\uparrow$} 
        & \makecell{\textbf{$\delta <$} \\ \textbf{$1.25^3$} $\uparrow$} 
        & \makecell{\textbf{Trainable} \\ \textbf{Attacker} $\downarrow$}
        & \makecell{\textbf{Heuristic} \\ \textbf{Score} $\uparrow$}
        \\
        \midrule
        
        Standard SemCom & \textbf{0.4186} & \textbf{0.7455} & \textbf{0.0213} & \textbf{70.71} & \textbf{93.49} & 1.0000 & 0.1206 \\
        
        Proposed ($\lambda_{\text{CTS}}=1$) & 0.3840 & 0.7416 & 0.0271 & 62.69 & 89.90 & \textbf{0.5612} & \textbf{0.3837} \\
        \bottomrule
    \end{tabular}
\end{table}

}

\subsection{Ablation Study}

We conduct an ablation study of our model to analyze the role of each component.
Specifically, we study the impact of contrastive learning and the training of the policy path.

\begin{table*}[ht]
	\scriptsize
    \centering
	\caption{Comparative Analysis of Different Ablation Models. }
    \label{table:cts_ablation}
	\setlength{\tabcolsep}{0.05cm}
	\begin{tabular}{c|cccc|ccccc|cc|cccc}
		\toprule 
		& \multicolumn{4}{c}{\makecell{\textbf{Semantic Segmentation}, $T_{\rm p}$}}
		& \multicolumn{5}{c}{\textbf{Depth Prediction}, $T_{\rm c}$ (\textbf{Stego Path})} 
		& \multicolumn{3}{c}{}
		
		\\\cmidrule(lr){2-5}\cmidrule(lr){6-10}
		\textbf{Model}
		& \multicolumn{2}{c}{\makecell{\textbf{Explicit Path}} }
		& \multicolumn{2}{c}{\makecell{\textbf{Stego Path}} }
		& \multicolumn{2}{c}{} & \multicolumn{3}{c}{$\mathbf{\delta < }$} 
		& \multicolumn{1}{c}{\textbf{Trainable}}
		& \multicolumn{1}{c}{\textbf{Heuristic}}
		& \multicolumn{4}{c}{\textbf{Relative Improvement (\%)}}  \\  
		\cmidrule(lr){2-3}
		\cmidrule(lr){4-5}
		\cmidrule(lr){8-10}
		\cmidrule(lr){13-16}
		
		& \makecell{\textbf{mIoU} $\uparrow$} 
		& \makecell{\textbf{Pixel}  \textbf{Acc} $\uparrow$} 
		
		& \makecell{\textbf{mIoU}  $\uparrow$} 
		& \makecell{\textbf{Pixel}  \textbf{Acc} $\uparrow$} 
		
		& \makecell{\textbf{abs}  \textbf{err} $\downarrow$} 
		& \makecell{\textbf{rel}  \textbf{err} $\downarrow$} 
		& \makecell{  $1.25 \uparrow$} 
		& \makecell{  $1.25^{2}\uparrow$} 
		& \makecell{ $ 1.25^{3}\uparrow$} 
		& \makecell{\textbf{Attacker $\downarrow$}}
		& \makecell{\textbf{Attacker $\uparrow$}}
		& \makecell{\textbf{rel}  \textbf{task}} 
		& \makecell{\textbf{rel}  \textbf{heu}} 
		& \makecell{\textbf{rel}  \textbf{tra}}
		& \makecell{\textbf{avg}  \textbf{all}} \\
		\midrule
        \color{black}Proposed ($\lambda_{\text{CTS}} = 2$) & \color{black}0.3569 & \color{black}0.7331 & \color{black}0.3534 & \color{black}0.7310 & \color{black}0.0308 & \color{black}0.4128 & \color{black}54.13 & \color{black}83.22 & \color{black}91.57 & \color{black}0.5325 & \color{black}0.3939 & \color{black}3.60&\color{black}-2.59 &\color{black}-5.39 &\color{black}-1.46 \\
		Proposed ($\lambda_{\text{CTS}} = 1$) & 0.3835 & 0.7403  & 0.3840 & 0.7416  & 0.0271 & 0.4598 & 62.69 & 81.69 & 89.90 &0.5612 & 0.3837& - & - & - &- \\

        non-CTS ($\lambda_{\text{CTS}} = 0$) & 0.3884 & 0.7406  & 0.3909 & 0.7412  & 0.0240 & 0.4353 & 65.92 & 83.33 & 91.07  & 0.7550 & 0.2060  & -3.05 & 86.26 &25.67  & 36.29\\

		Similarity & 0.3743 & 0.7380  & 0.3755 & 0.7374  & 0.0362 & 0.5774 & 43.79 & 77.89 & 86.69 &0.9325 & 0.1533 &10.43 & 150.3   & 39.82 & 66.85\\
		Random Path & 0.3591& 0.7342  & 0.3630 & 0.7346  & 0.0271 & 0.4586 & 62.03 & 82.09 & 89.96  & 0.6530 &0.3720 &  1.82& 3.15  & 14.06&6.34 \\
		\bottomrule
  \multicolumn{16}{l}{
\makecell[tl]{
\textit{``rel task''} denotes the total relative enhancement across all task metrics; 
\textit{``rel heu''} denotes the relative improvement in Heuristic Attacker value;\\
\textit{``rel tra''} denotes the relative improvement in Trainable Attacker value;\\
\textit{``avg all''} is the mean of \textit{rel task}, \textit{rel heu}, and \textit{rel tra}.\\
\textit{Relative improvement (RI) is computed following the arrow directions:} 
for metrics with $\uparrow$, $\mathrm{RI}=\frac{\mathrm{Proposed}-\mathrm{Baseline}}{\mathrm{Baseline}}$; 
for metrics with $\downarrow$, $\mathrm{RI}=\frac{\mathrm{Baseline}-\mathrm{Proposed}}{\mathrm{Baseline}}$.\\
}
}

	\end{tabular}
	
\end{table*}

\begin{figure}[t]
	\centering
    \includegraphics[width=0.5\textwidth]{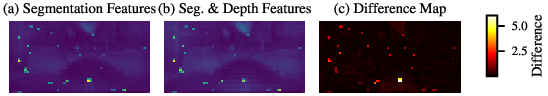}
	\caption{Feature distribution of training with a random policy path.}
	\label{fig:random_path}
\end{figure}

\begin{figure}[t]
	\centering
    \includegraphics[width=0.5\textwidth]{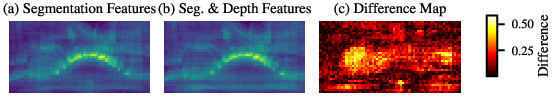}
	\caption{Feature distribution of training with maximizing cosine similarity.}
	\label{fig:cos}
\end{figure}

{\color{black}
\begin{figure}[htbp]
	\centering
	\includegraphics[width=0.8\linewidth]{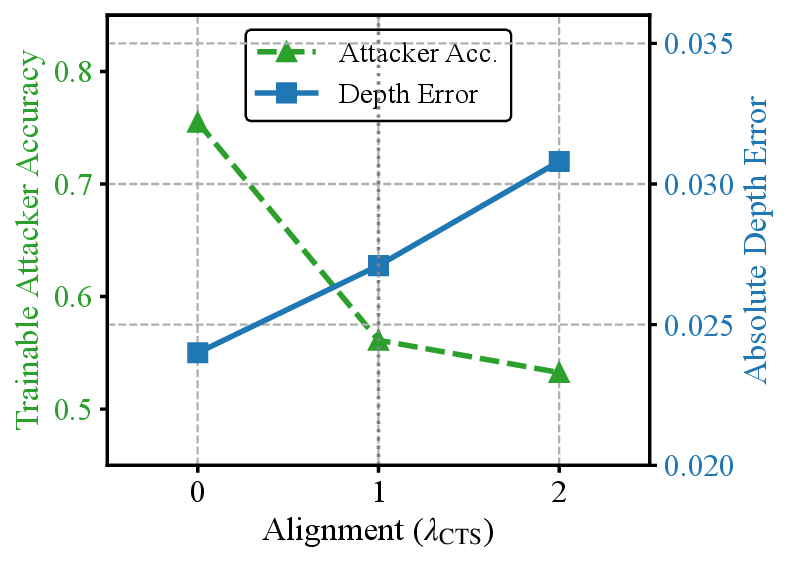}
	\caption{Trade-off analysis under varying weighting factors $\lambda_{\text{CTS}}$.}
	\label{fig:cts_tradeoff}
\end{figure}

\vspace{1em} 

\begin{figure}[htbp]
	\centering
	\includegraphics[width=0.8\linewidth]{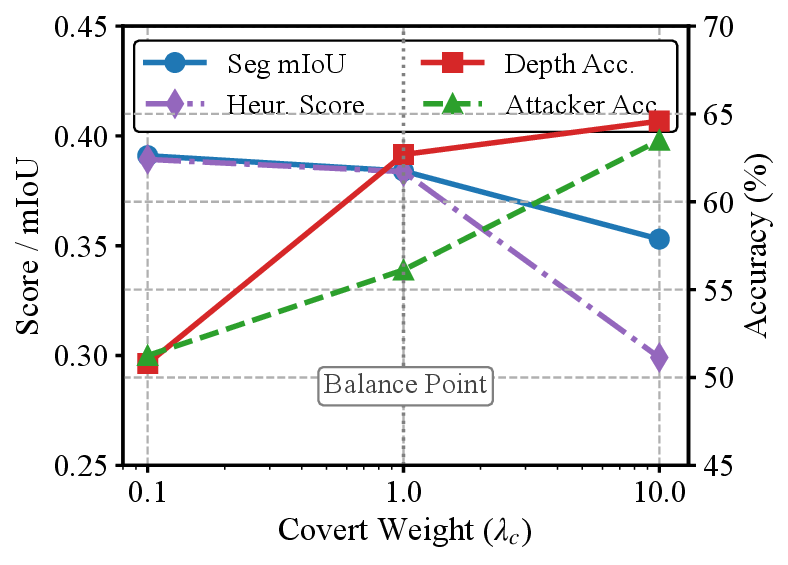}
	\caption{Trade-off analysis between public and covert tasks under varying $\lambda_{\text{c}}$.}
	\label{fig:lambda_tradeoff_response}
\end{figure}

\subsubsection{Impact of Contrastive Learning and Trade-off Analysis}
\label{sec:impact_of_cts}

To explicitly address the trade-off between feature-level indistinguishability and covert task capacity, we analyze the impact of the contrastive alignment mechanism by varying its weighting factor $\lambda_{\text{CTS}} \in \{0, 1, 2\}$.
The quantitative results are listed in Table~\ref{table:cts_ablation}, and the trade-off trends are visualized in Fig.~\ref{fig:cts_tradeoff}.

In the absence of the alignment constraint ($\lambda_{\text{CTS}} = 0$), the Stego path is free to optimize the feature representation solely for task performance.
Consequently, this unconstrained baseline achieves the lowest \textit{abs err} on the covert depth prediction task (\textit{abs err}: 0.0240) alongside high accuracy on the public segmentation task. 
However, without the constraint to mimic the public features, the Stego features remain statistically distinct, rendering the system highly vulnerable to detection with an attacker accuracy of 75.50\%.

Increasing the weighting factor to $\lambda_{\text{CTS}} = 2$ enforces a strict similarity constraint, highlighting a pronounced trade-off between security and capacity.
On the one hand, the trainable attacker detection accuracy drops significantly to 53.25\%, close to random guessing and indicating near-perfect feature indistinguishability. On the other hand, this strict alignment implicitly limits the feature variance available for encoding the covert signal. This over-regularization leads to a noticeable degradation in task performance, where the depth \textit{abs err} rises to 0.0308. Furthermore, the public segmentation performance also declines (\textit{mIoU} drops to 0.3569), suggesting that excessive alignment constraints create optimization conflicts that impair the learning of semantic representations.

Based on this analysis, we adopt $\lambda_{\text{CTS}} = 1$ as the optimal operating point within this trade-off.
It leverages the redundancy in the feature space to hide the covert signal effectively, suppressing the trainable attacker detection accuracy to 56.12\% with only a moderate increase in task \textit{abs err} (0.0271) compared to the unconstrained baseline.
This analysis confirms that $\lambda_{\text{CTS}}$ serves as a pivotal hyperparameter. While stronger alignment improves stealth, it inevitably constrains the covert rate. Our framework effectively navigates this trade-off, as evidenced by a substantial 36.29\% relative improvement in the comprehensive \textit{avg all} metric over the non-CTS baseline, thereby securing the communication channel with a manageable cost to utility.

}

\subsubsection{Substituting CTS with Cosine Similarity}

In our ablation study, we replaced the CTS mechanism with a simpler objective: directly maximizing the cosine similarity between feature pairs. The results, presented in Table~\ref{table:cts_ablation}, reveal that this seemingly intuitive approach leads to a significant degradation in both task performance and, most critically, security.

While the Cosine Similarity method performs comparably to our full model on the public segmentation task, its performance on the covert Depth Prediction task collapses. This is evidenced by the $\delta < 1.25$ metric dropping by 30.1\%, indicating that the features generated under this objective lose task-critical details.

The security implications of this substitution are even more stark. Our adversarial evaluation reveals a catastrophic failure, with the trainable attacker achieving a near-perfect detection accuracy of 93.25\% against the Cosine Similarity model. This stands in extreme contrast to the 56.12\% accuracy achieved against our proposed CTS-based model. This security vulnerability is also reflected in the internal heuristic score, which plummets from 0.3837 to 0.1533.

The fundamental reason for this discrepancy is that maximizing cosine similarity alone encourages a trivial solution known as feature space collapse. The model learns to map all inputs to a similar, low-variance region of the feature space, creating an unnatural distribution that is trivial for an adversary to detect. Conversely, our CTS objective, by jointly optimizing positive and negative sample pairs, acts as a powerful regularizer. This push-and-pull mechanism not only aligns the features for covertness but also preserves a rich and well-structured feature space, which is essential for both robust task performance and verifiable security.

\subsubsection{Impact of Adaptive Path Selection}
\label{sec:impact_of_path_selection}

To isolate the contribution of our learnable policy, we compare our full model against a Random Path baseline, where the encoder path is selected randomly without training optimization. Table~\ref{table:cts_ablation} demonstrates the clear superiority of our policy-driven approach on both security and task performance fronts.

In terms of security, our learnable policy is substantially more effective. It reduces the trainable attacker's detection accuracy to a near-random 56.12\%, whereas the Random Path model is significantly more vulnerable at 65.30\% accuracy. This confirms that a random policy, lacking explicit optimization by the CTS loss, leaves behind more easily exploitable statistical artifacts.

Concurrently, this security gain is achieved alongside better utility. Our policy-driven model achieves a 1.82\% relative improvement in average task performance compared to the random path model. This dual improvement highlights that a learnable policy is critical: it learns to find paths that are not only more effective for the given tasks by focusing on salient features, but are also optimized to be statistically stealthier.

\begin{figure}[t]
	\centering
    \includegraphics[width=0.45\textwidth]{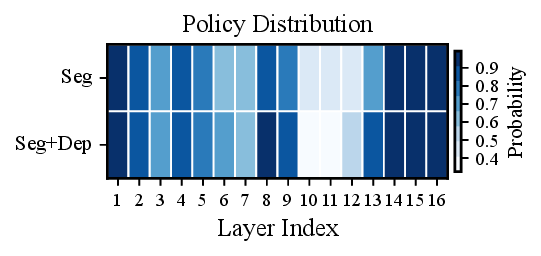}
	\caption{Visualization of the Select-or-Skip policy in the encoder blocks, i.e., the policy distribution $\{\alpha_l^k \mid l = 1, \dots, 16,\, k \in \{\mathrm{Exp}, \mathrm{Ste}\}\}$. Comparisons are made between the Explicit path (Seg) and the Stego path (Seg+Dep).
	}
	\label{fig:policy_vis}
\end{figure}

Fig.~\ref{fig:policy_vis} visualizes the select-or-skip policy distributions of Explicit Path (Seg) versus Stego Path (Seg+Dep) through block-wise activation probabilities (ResNet-34 has 16 blocks).
The heatmap reveals differential sharing patterns across network depth.
In early blocks (1 -- 6), both policies exhibit high sharing rates ($>75\%$). The middle blocks (7 -- 12) diverge significantly: The Explicit Path maintains $45\%$--$62\%$ sharing in mid-blocks, while the Stego Path drops to $32\%$--$51\%$ sharing; covert tasks require dedicated feature processing in mid-level abstraction blocks where task-specific patterns emerge.
These patterns align with feature abstraction hierarchies. Low-level features (edges, textures) exhibit universal utility. High-level representations require task-specific adaptation.

\subsection{Impact of Hyperparameters}
\label{sec:impact_of_hyperparameters}

\subsubsection{Sparsity-Aware Regularization ($\beta$ and $\gamma$)}

\begin{table*}[ht]
	\scriptsize
    \centering
	\caption{Ablation Study on the Sparsity-Aware Regularization Hyperparameters ($\beta$ and $\gamma$)}
    \label{table:hyper_analy}
	\setlength{\tabcolsep}{0.1cm}
	\begin{tabular}{c|cccc|ccccc|c|ccccc}
		\toprule 
		& \multicolumn{4}{c}{\makecell{\textbf{Semantic Segmentation}, $T_{\rm p}$}}
		& \multicolumn{5}{c}{\textbf{Depth Prediction}, $T_{\rm c}$ (\textbf{Stego Path})} 
		& \multicolumn{5}{c}{}
		
		\\\cmidrule(lr){2-5}\cmidrule(lr){6-10}
		\textbf{Model}
		& \multicolumn{2}{c}{\makecell{\textbf{Explicit Path}} }
		& \multicolumn{2}{c}{\makecell{\textbf{Stego Path}} }
		& \multicolumn{2}{c}{} & \multicolumn{3}{c}{$\mathbf{\delta < }$} 
		& \multicolumn{1}{c}{\textbf{Attacker}}
		& \multicolumn{4}{c}{\textbf{Relative Improvement (\%)}}  \\  
		\cmidrule(lr){2-3}
		\cmidrule(lr){4-5}
		\cmidrule(lr){8-10}
		\cmidrule(lr){12-15}
		
		& \makecell{\textbf{mIoU} $\uparrow$} 
		& \makecell{\textbf{Pixel}  \textbf{Acc} $\uparrow$} 
		
		& \makecell{\textbf{mIoU}  $\uparrow$} 
		& \makecell{\textbf{Pixel}  \textbf{Acc} $\uparrow$} 
		
		& \makecell{\textbf{abs}  \textbf{err} $\downarrow$} 
		& \makecell{\textbf{rel}  \textbf{err} $\downarrow$} 
		& \makecell{  $1.25 \uparrow$} 
		& \makecell{  $1.25^{2}\uparrow$} 
		& \makecell{ $ 1.25^{3}\uparrow$}
		& \makecell{\textbf{Acc}}
		& \makecell{\textbf{rel}  \textbf{task}} 
		& \makecell{\textbf{rel}  \textbf{heu}}
		& \makecell{\textbf{rel}  \textbf{tra}}
		& \makecell{\textbf{avg}  \textbf{all}} \\
		\midrule
		Proposed  & \textbf{ 0.3830} & \textbf{0.7403}  & \textbf{0.3849} & \textbf{0.7416}  & \textbf{0.0271} & 0.4597 & \textbf{62.66} & 81.69 & 89.90 & \textbf{0.5612} & - & - & - &- \\
		$\beta = 0$ & 0.3726 & 0.7375  & 0.3730 &  0.7377&  0.0289 & 0.4513  & 58.97 & 82.61 &90.44 &0.6350 & 1.74 & 5.21 & 11.62 &  6.19\\
		$\gamma = 0$  & 0.3708 & 0.7342  & 0.3716 & 0.7355  & 0.0292 & \textbf{0.4242} & 58.43 & \textbf{83.49} & \textbf{91.34} & 0.6312 & 1.33 & 2.95 &11.09& 5.12\\
		\bottomrule
	\end{tabular}

\end{table*}

We validate the necessity of the two sparsity-aware regularization terms, controlled by $\beta$ and $\gamma$, by analyzing their impact on both task performance and the system's resilience against a trainable attacker. As shown in Table~\ref{table:hyper_analy}, disabling either term leads to a significant degradation in security, confirming their distinct yet complementary roles in achieving a robust balance between utility and stealth.

\textbf{Impact of Sparsity Control ($\beta$):}
The $\ell_1$-norm sparsity penalty, controlled by $\beta$, encourages the model to select fewer, more essential encoder blocks. Disabling this term ($\beta = 0$) forces the network to use a denser, more complex path. This measurably degrades security, increasing the attacker's detection accuracy from our model's baseline of 56.12\% to 63.50\%. The reason is that enforcing sparsity compels the model to generate a more compact and less complex feature representation. A denser path, conversely, creates richer and more intricate features, which inadvertently provides a larger attack surface with more unique statistical artifacts for an adversary to learn.
{\color{black}This contribution is quantitatively confirmed by the comprehensive \textit{avg all} metric, where the proposed sparsity mechanism yields a 6.19\% relative improvement over the non-sparse baseline.}

\textbf{Impact of Path Distribution Alignment ($\gamma$):}
This term imposes a layer-weighted penalty on deviations between the Stego Path and Explicit Path selection distributions, forcing the encoder to share more blocks in the early layers while allowing for more divergence in the later layers. This design is crucial: it aligns the paths where they learn generic, low-level features (early layers), {\color{black} ensuring the sharing of foundational information essential for task utility,} while permitting the necessary divergence for learning distinct, task-specific features (later layers). Removing this constraint ($\gamma = 0$) results in {\color{black} a simultaneous degradation in both system utility and security. Specifically, the public segmentation \textit{mIoU} drops to 0.3708, and} the attacker's accuracy surges to 63.12\%. This occurs because the Stego Path is free to diverge structurally from the outset, {\color{black} losing the robust shared features required for maintaining task precision and} creating a blatant and easily exploitable statistical discrepancy.

{\color{black}
\subsubsection{Impact of Covert Task Weight $\lambda_{\text{c}}$}

To investigate the inherent trade-off between public task performance and covert transmission capacity, we conduct a sensitivity analysis on the covert task balancing coefficient, $\lambda_{\text{c}}$. We evaluate the model performance under varying weights $\lambda_{\text{c}} \in \{0.1, 1.0, 10\}$.

The results, illustrated in Fig. \ref{fig:lambda_tradeoff_response}, reveal a distinct conflict between the two objectives within the Stego path. When the covert task is prioritized ($\lambda_{\text{c}}=10$), the system achieves superior depth prediction accuracy ($\delta_{1.25}$ reaches 64.6). However, this suppresses the optimal expression of public semantics, causing the  \textit{mIoU} to drop to 0.353. Consequently, the identifiable feature artifacts become easily detectable by the attacker (detection accuracy rises to 63.5\%).

Conversely, reducing the covert weight ($\lambda_{\text{c}}=0.1$) allows the public task to dominate feature generation. The Stego features align closely with the explicit path, ensuring high cover quality (Segmentation \textit{mIoU} of 0.393) and excellent indistinguishability (trainable attacker accuracy drops to 51.25\%). However, this comes at the cost of the covert task, where the depth estimation performance degrades significantly ($\delta_{1.25}$ falls to 50.8). Based on these observations, we adopt $\lambda_{\text{c}}=1.0$ as the optimal setting to balance task performance and security.}

\section{Conclusion}\label{sec:conclusion} 

{\color{black}
	
	In this paper, we have proposed a novel adaptive dual-path framework for covert semantic communication, embedding a hidden task directly into the semantic features of a public one. Our dual-path adaptive architecture, guided by a security-driven contrastive learning objective, has rendered the public and steganographic feature streams statistically indistinguishable. A rigorous adversarial evaluation has validated our approach: the framework has suppressed a powerful attacker’s detection accuracy to a near-random 56.12\%, while baselines have been detected with approximately 100\% accuracy. This state-of-the-art security has been achieved while maintaining superior task performance.
	
	This work has established an effective paradigm for secure SemCom by shifting the embedding process to the feature space and validating it against a learning-based adversary. It has demonstrated a viable path toward high-capacity, provably covert systems and has revealed novel security–utility trade-offs. {\color{black} 
    Looking forward, a promising direction is to extend the framework to dynamic environments characterized by time-varying links.
    Inspired by recent advances in multilevel feature transmission~\cite{yan2023multilevel}, we envision integrating an online transmission decision module (e.g., based on Deep Reinforcement Learning) atop the current architecture. }
    {\color{black}
A natural extension is to cast the transmission scheduling under time-varying links as a constrained Markov decision process with a delay-threshold constraint, while maintaining the covertness constraint in the semantic feature space.}
    	{\color{black}
    Unlike the current internal topology search, this agent would operate at the transmission interface to handle Markovian channel evolution.
    It would dynamically adjust the transmitted information volume and granularity based on real-time channel states and delay constraints, thereby adaptively balancing utility, security, and latency in fluctuating environments.}

{\color{black}
Another important future direction is to further extend the current same-source coupled framework to more general non-aligned scenarios, where the covert target may originate from a different image source. In such cases, the shared spatial and semantic correspondence exploited by the present design is no longer directly available in the same form, which calls for a more general cross-source semantic integration mechanism. A promising extension is to decouple the initial feature extraction by employing source-specific front-end encoders, followed by a latent-level hiding/fusion module to integrate heterogeneous semantic representations before they are processed by the adaptive dual-path shared backbone. In this way, source heterogeneity can be handled at the front end while preserving the core alignment and indistinguishability mechanisms of the proposed framework.
}}

\appendices 
 
\section{Convergence Analysis}
\label{sec:appendix_convergence}

\begin{figure*}[t] 
	\centering
	
	\subfloat[Public Task]{
		\includegraphics[width=0.3\textwidth]{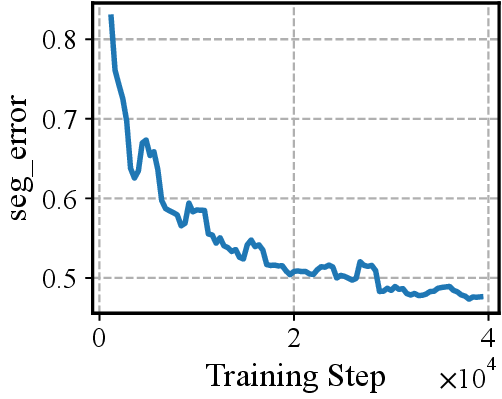}
		\label{fig:conv_seg_proposed}
	}
	\hfill 
	\subfloat[Public Task (Stego)]{
		\includegraphics[width=0.3\textwidth]{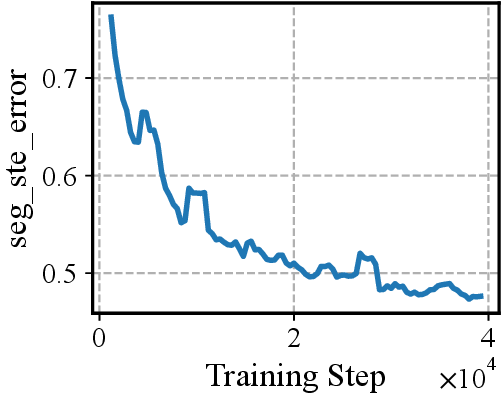}
		\label{fig:conv_seg_ste_proposed}
	}
	\hfill 
	\subfloat[Covert Task]{
		\includegraphics[width=0.3\textwidth]{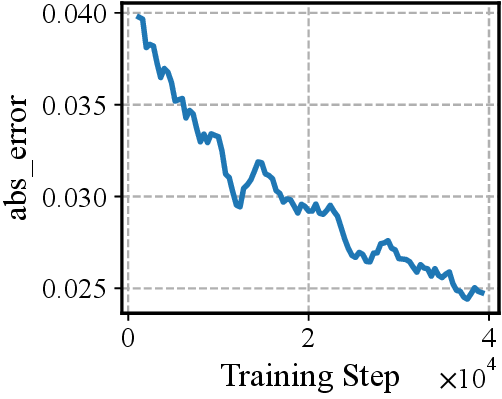}
		\label{fig:conv_abs_proposed}
	}
	
	\caption{
		Training convergence analysis. The $x$-axis represents training steps, and the $y$-axis represents the task error.
		\textbf{(a)} Segmentation error on the Explicit Path (Public Baseline). 
		\textbf{(b)} Segmentation error on the Stego Path, showing that embedding covert data has minimal impact on public task convergence. 
		\textbf{(c)} Absolute error of depth prediction (Covert Task), demonstrating stable convergence.
	}
	\label{fig:convergence_curves_proposed}
\end{figure*}

The stability of our multi-objective training framework is critical for ensuring reliable performance. Fig.~\ref{fig:convergence_curves_proposed} visualizes the training error curves, showing a consistent and smooth decrease for all tasks. This result validates the stable convergence of our model and confirms that the competing objectives are effectively balanced during training.

{\color{black}
\section{Complexity Analysis}
\label{sec:appendix_complexity}
To demonstrate the efficiency advantages of our unified architecture, we compared the proposed framework against standard baselines, specifically the Stacking-based method (Base Model 1) and the Noise-based method (Base Model 2).
As detailed in Table \ref{tab:complexity_comparison}, our method reduces the parameter count by exactly 50\% (21.30 M vs. 42.60 M) compared to Base Model 1, attributed to the unified shared backbone architecture.
Moreover, despite executing two functional paths (Explicit and Stego), the adaptive block-skipping mechanism ensures that each path is sparse. This leads to a 13.4\% reduction in total FLOPs compared to the fully dense baseline.
While the inference latency (19.18 ms) is marginally higher than Base Model 1 (17.83 ms) due to the gating operations, it remains competitive and faster than the Noise-based Base Model 2 (19.48 ms). 
This trade-off yields substantial gains in storage and computational efficiency, making the proposed framework highly suitable for efficient edge deployment.

\begin{table}[H] 
\centering
\color{black}
\caption{Complexity Comparison with Baselines}
\label{tab:complexity_comparison}
\renewcommand{\arraystretch}{1.2}
\setlength{\tabcolsep}{2.5pt} %
\small
\begin{tabular}{l|c|c|c}
\toprule
\textbf{Method} & \textbf{Params (M)} $\downarrow$ & \textbf{FLOPs (G)} $\downarrow$ & \textbf{Avg. Time (ms)} $\downarrow$ \\
\midrule
Base Model 1 & 42.60 & 37.06 & \textbf{17.83} \\
Base Model 2 & 42.61 & 37.06 & 19.48 \\
\textbf{Proposed Model} & \textbf{21.30} & \textbf{32.10} & 19.18 \\
\bottomrule
\end{tabular}
\end{table}

}

\bibliographystyle{IEEEtran}

\bibliography{citations}

    \begin{IEEEbiography}[{\includegraphics[width=1in,height=1.25in,clip,keepaspectratio]{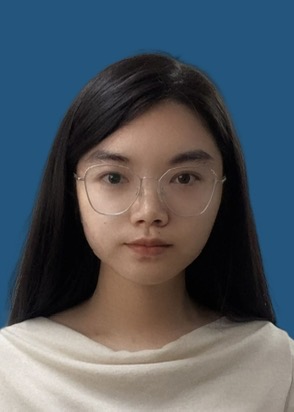}}]{Xi Yu} (Graduate Student Member, IEEE)
received the B.E. degree in communication engineering from Beijing University of Posts and Telecommunications (BUPT), China, in 2020. She is pursuing her Ph.D. with the School of Information and Communication Engineering at BUPT. Her research interests include multi-task semantic communication and privacy-preserving techniques.
\end{IEEEbiography}\vspace{-20 mm}

\begin{IEEEbiography}[{\includegraphics[width=1in,height=1.25in,clip,keepaspectratio]{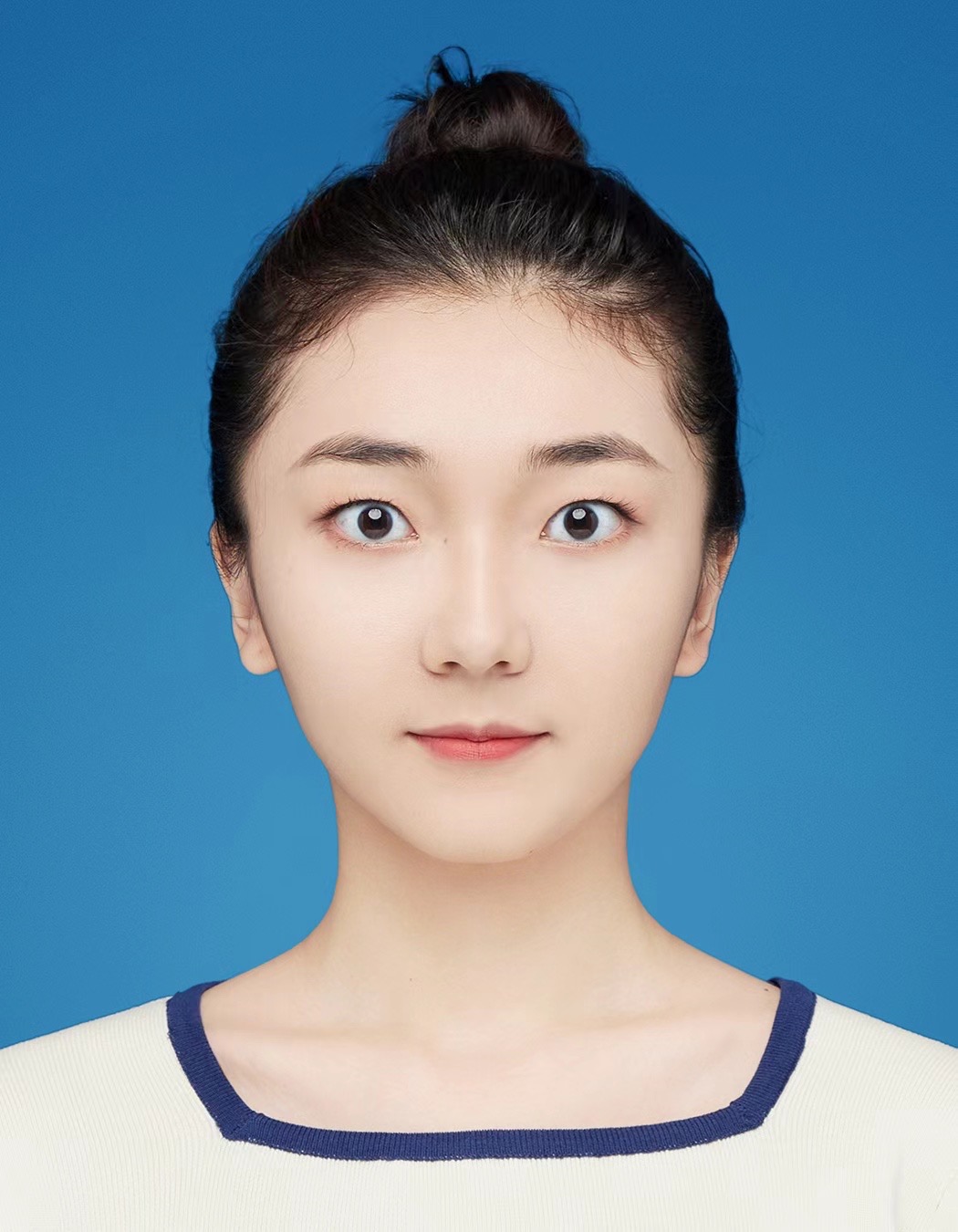}}]{Weicai Li} (Graduate Student Member, IEEE)
received the B.E. degree in communication engineering from Beijing University of Posts and Telecommunications (BUPT), China, in 2020. She is pursuing her Ph.D. with the School of Information and Communication Engineering at BUPT. From December 2022 to December 2023, she was a Visiting Scholar at the University of Technology Sydney. Her research interests include wireless federated learning, distributed computing, and privacy-preserving.
\end{IEEEbiography}\vspace{-10 mm}

\begin{IEEEbiography}[{\includegraphics[width=1in,height=1.25in,clip,keepaspectratio]{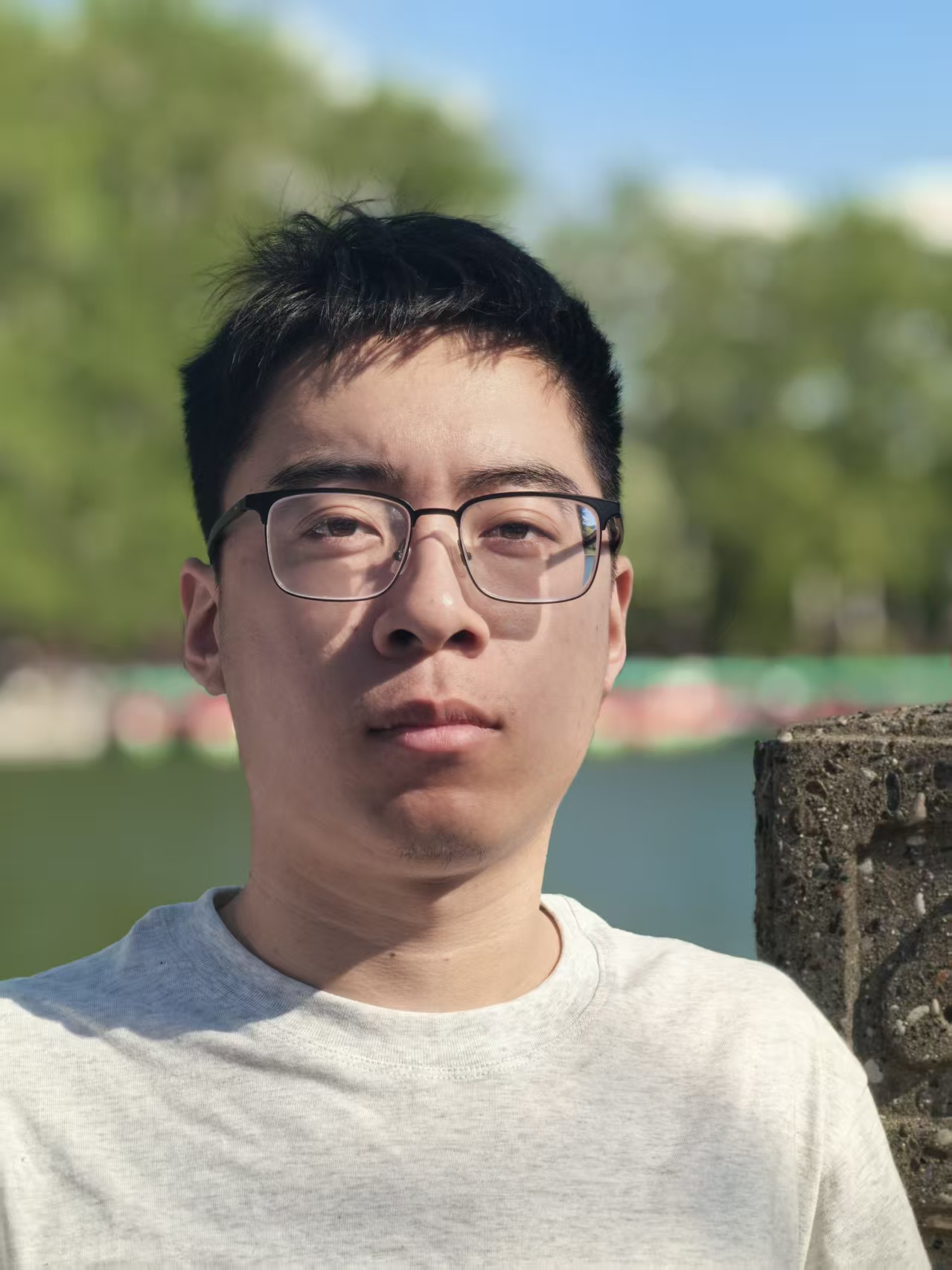}}]{Lin Yin} 
received the B.E. degree in applied physics from Beijing University of Posts and Telecommunications (BUPT), China, in 2023. He is pursuing his Ph.D. with the School of Information and Communication Engineering at BUPT. His research interests include personalized federated learning and semantic communication .
\end{IEEEbiography}\vspace{-20 mm}

\begin{IEEEbiography}[{\includegraphics[width=1in,height=1.25in,clip,keepaspectratio]{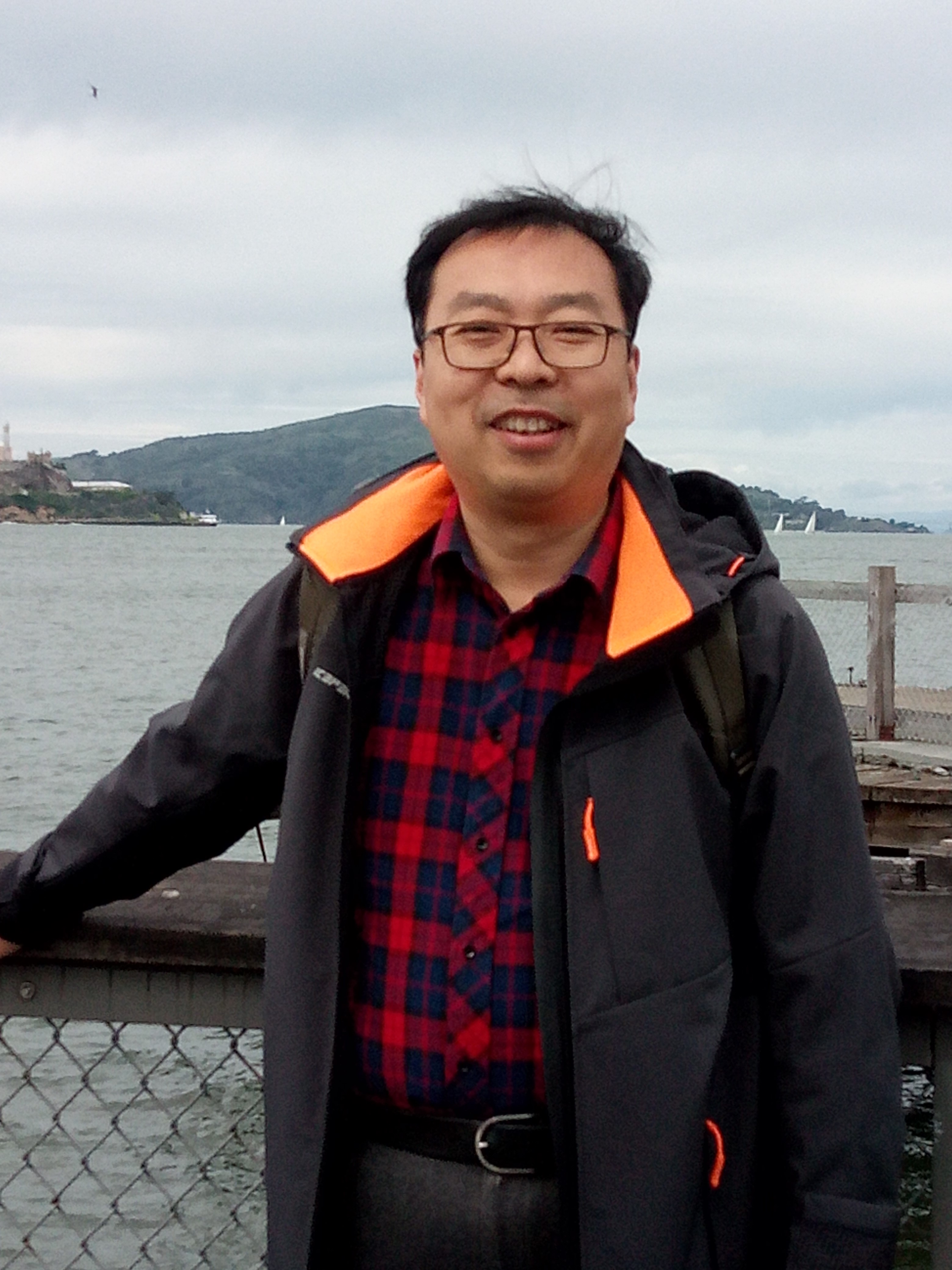}}]{Tiejun Lv} received the M.S. and Ph.D. degrees in electronic engineering from the University of Electronic Science and Technology of China (UESTC), Chengdu, China, in 1997 and 2000, respectively. From January 2001 to January 2003, he was a Post-Doctoral Fellow at Tsinghua University, Beijing, China. In 2005, he was promoted to a Full Professor at the School of Information and Communication Engineering, Beijing University of Posts and Telecommunications (BUPT). From September 2008 to March 2009, he was a Visiting Professor with the Department of Electrical Engineering, Stanford University, Stanford, CA, USA. He is currently the author of four books, one book chapter, and more than 170 published journal articles and 200 conference papers on the physical layer of wireless mobile communications. His current research interests include signal processing, communications theory, and networking. He was a recipient of the Program for New Century Excellent Talents in University Award from the Ministry of Education, China, in 2006. He received the Nature Science Award from the Ministry of Education of China for the hierarchical cooperative communication theory and technologies in 2015 and Shaanxi Higher Education Institutions Outstanding Scientific Research Achievement Award in 2025.

\end{IEEEbiography}\vspace{-10 mm}

\end{document}